\documentclass[sigconf]{acmart}
\AtBeginDocument{%
  }
    
\usepackage{algorithm}
\usepackage{algpseudocode}
\usepackage{subcaption}
\usepackage{tabularx}
\usepackage{multirow}     
\usepackage{booktabs}     
\usepackage{amsmath}
\usepackage{float}

\setcopyright{acmlicensed}
\copyrightyear{2018}
\acmYear{2018}
\acmDOI{XXXXXXX.XXXXXXX}
\acmConference[Conference acronym 'XX]{Make sure to enter the correct
  conference title from your rights confirmation email}{June 03--05,
  2018}{Woodstock, NY}
\acmISBN{978-1-4503-XXXX-X/2018/06}

\begin{document}

\title{FedGA: A Fair Federated Learning Framework Based on the Gini Coefficient}
\subtitle{Research}
\author{ShanBin Liu}
\email{shanbinliu0@gmail.com}
\affiliation{%
  \institution{Guangxi Key Laboratory of Trusted Software, Guilin University of Electronic Technology}
  \city{Guilin}
  \state{Guangxi}
  \country{China}
}









\begin{abstract}
  Fairness has emerged as one of the key challenges in federated learning. In horizontal federated settings, data heterogeneity often leads to substantial performance disparities across clients, raising concerns about equitable model behavior. To address this issue, we propose FedGA, a fairness-aware federated learning algorithm. We first employ the Gini coefficient to measure the performance disparity among clients. Based on this, we establish a relationship between the Gini coefficient $G$ and the update scale of the global model ${U_s}$, and use this relationship to adaptively determine the timing of fairness intervention. Subsequently, we dynamically adjust the aggregation weights according to the system’s real-time fairness status, enabling the global model to better incorporate information from clients with relatively poor performance.We conduct extensive experiments on the Office-Caltech-10, CIFAR-10, and Synthetic datasets. The results show that FedGA effectively improves fairness metrics such as variance and the Gini coefficient, while maintaining strong overall performance, demonstrating the effectiveness of our approach.
\end{abstract}




\maketitle

\section{Introduction}
The widespread use of connected devices generates vast distributed data crucial for training AI models.  However, traditional centralized training raises privacy, ownership, and regulatory concerns, while single-device training suffers from limited data. These challenges motivated federated learning \cite{r2}, which enables collaborative training without raw data transfer, preserving privacy while achieving comparable performance to centralized methods.\\
Yet federated learning faces fairness challenges due to heterogeneous client data distributions. Performance disparities across clients can undermine trust, participation, and model reliability—particularly critical in healthcare and finance where equitable performance is essential.\\
To address this, we propose FedGA, a fairness-aware federated learning algorithm that monitors Gini coefficient dynamics to determine optimal fairness intervention timing and adaptively adjusts optimization intensity based on client validation performance.\\
The main contributions of this work are as follows:
\begin{itemize}
    \item We investigate the relationship between the Gini coefficient and the global update scale  during federated learning, and observe that they tend to decrease concurrently. Based on this observation, we propose a novel delayed fairness intervention strategy.
    \item We design an algorithm that dynamically adjusts aggregation weights based on client validation set performance. Additionally, we introduce a hyperparameter $\lambda$ to control the degree of fairness intervention, enabling practitioners to flexibly balance fairness and accuracy according to specific requirements.
    \item We provide a theoretical guarantee that the aggregation weight of the best-performing client is always less than $\frac{1}{n}$, while the weight of the worst-performing client is always greater than $\frac{1}{n}$, where $n$ denotes the number of participating clients in each communication round.
    \item We analyze the time complexity of the delayed fairness intervention strategy. Compared to FedGini, our method FedGA exhibits lower computational complexity when the number of clients is smaller than the number of model parameters.
    \item We theoretically establish the relationship between the Gini coefficient and average sum of accuracy differences among clients (denoted as AvgDiff) during the later stages of federated learning training, where the global model has largely stabilized. Specifically, we derive a first-order approximation showing that changes in Gini and mean accuracy jointly influence AvgDiff, thereby providing a formal link between fairness metrics and client-level performance consistency.
    \item We conduct extensive experiments on two real-world datasets and one synthetic dataset. Specifically, feature shift and label shift are simulated on the Office-Caltech-10 and CIFAR-10 datasets to represent two distinct types of heterogeneity. The experimental results validate the effectiveness of the proposed method in improving both fairness and performance.
\end{itemize}

\section{Background And Motivation}
In this section, we first introduce the optimization objectives of federated learning in Section 2.1. Then, in Sections 2.2 and 2.3, we present the definitions and evaluation metrics of fairness in federated learning. Finally, in Section 2.4, we discuss our motivation-how to search for an appropriate timing for fairness intervention while keeping the computational overhead minimal.


\subsection{Optimization Goals of Federated Learning}
The federated learning process consists of three iterative steps: (1) server distributes the global model to selected clients; (2) clients train locally on private data and upload updated models; (3) server aggregates client models to form a new global model.These steps continue until convergence or a predefined number of rounds. The optimization objective is:
\begin{equation}
    \mathop {\min }\limits_w f\left( w \right) = \mathop \sum \limits_{k = 1}^m {p_k}{F_k}\left( w \right)
\end{equation}
\begin{equation}
    {F_k}(w) = \frac{1}{{{n_k}}}\sum\limits_{{j_{k = 1}}}^{{n_k}} {{l_{{j_k}}}(w)}
\end{equation}
The formal description of this objective is as follows:
\begin{equation}
    \mathop {\min }\limits_w f(w) = \left\langle {w,m,k,{P_k},{F_k}(w),{n_k},{j_k},{l_{{j_k}}}(w)} \right\rangle
\end{equation}
Where:\\
1. $f(w)$ is the global optimization objective of federated learning.\\
2. $w$ is the global model of federated learning.\\
3. $m$ is the total number of clients participating in this round of training.\\
4. $k$ is the index of the client.\\
5. ${P_k}$ is the aggregation weight of client $k$.${P_k} \ge 0$ and $\sum\limits_{k = 1}^m {{p_k}}  = 1$.Typically, ${p_k} = \frac{{{n_k}}}{n}$ or ${p_k} = \frac{1}{m}$, where $n$ is the total size of the dataset owned by all devices participating in this round of federated learning.\\
6. ${F_k}$ is the local optimization objective of client $k$.\\
7. ${n_k}$ is the amount of data owned by client $k$.\\
8. ${j_k}$ is the index of a data sample.\\
9. ${l_{{j_k}}}(w)$ is the loss function of the global model parameter $w$ in sample ${j_k}$.



\subsection{Definition of Fairness In Federated Learning And Metrics For Measuring Fairness}
Our definition of fairness follows Li et al. \cite{r3}.For two models ${{\rm{w}}_1}$ and ${{\rm{w}}_2}$, if the performance distribution $\left\{ {{\rm{a}}_1^{{w_1}},...,{\rm{a}}_n^{{w_1}}} \right\}$ of model ${{\rm{w}}_1}$ is more uniform than the performance distribution $\left\{ {{\rm{a}}_1^{{w_2}},...,{\rm{a}}_n^{{w_2}}} \right\}$ of model ${{\rm{w}}_2}$ , then model ${{\rm{w}}_1}$ is considered to be fairer than ${{\rm{w}}_2}$ . Here,  $a_i^w$ denotes the performance of model $w$ on client $i$, which can be either accuracy or loss. In this work, We adopt standard deviation and Gini coefficient as fairness metrics, where lower values indicate more uniform client performance distribution and thus fairer federated learning outcomes.



\subsection{Gini Coefficient}
The Gini coefficient \cite{r35}, proposed by Corrado Gini based on the Lorenz curve, was originally designed to measure wealth inequality on a scale from 0 (perfect equality) to 1 (maximal inequality). In federated learning, it quantifies client performance imbalance: a value of 0 indicates identical performance across all clients (perfect fairness), while 1 represents extreme unfairness where only one client benefits from the global model. The formal definition of the Gini coefficient is given as follows:
\begin{equation}
    G = \frac{{\sum\limits_{i = 1}^n {\sum\limits_{j = 1}^n {\left| {{x_i} - {x_j}} \right|} } }}{{2\left( {{\rm{n - }}1} \right)\sum\limits_{j = 1}^n {{x_j}} }}
    \label{eq:gini}
\end{equation}
where ${x_i}$ denotes the accuracy of client $i$, and $n$ denotes the total number of clients.
\subsection{Motivation}
Most existing fairness optimization algorithms initiate intervention from the early stages of federated learning, which may inadvertently undermine fairness\cite{r25}. To mitigate this, Li et al. \cite{r25} proposed FedGini, which adaptively determines the intervention timing by monitoring the global update scale ${U_s}$ . While effective in avoiding premature intervention, this method incurs high computational complexity.\\
To address this limitation, we propose FedGA, a lightweight alternative that preserves adaptive fairness scheduling with significantly reduced overhead. As shown in Section 4.2, the time complexity of FedGini is $O(p \times q \times n)$, where $p \times q$ is the number of model parameters, and  $n$ is the number of clients participating in each round. This complexity increases with the size of the model and client population, which may present practical challenges in large-scale deployments. With the emergence of large language models (LLMs) such as ChatGPT and Claude, which contain hundreds of millions of parameters, the computational burden becomes especially pronounced. By contrast, FedGA reduces the time complexity to $O\left( {{n^2}} \right)$ , where $n$ denotes the number of clients per round. This design offers improved scalability and training efficiency, making FedGA better suited for large-scale federated learning scenarios.
\section{The Design of FedGA}
FedGA comprises two main components: a delayed fairness intervention strategy and dynamic adjustment of aggregation weights.
\subsection{Delayed Fairness Intervention Strategy Based on Gini Coefficient Aware}
Geyer et al. \cite{r36} proposed two definitions: the update scale ${{U}_{s}}$ and the sum over all parameter variances in the update matrix ${{V}_{c}}$.\\
\textbf{Definition 1:} The update scale${{U}_{s}}$. Let$\Delta {{w}_{i,j}}$define the $(i,j)th$ parameter in an update of the form $\Delta w\in {{R}^{p\times q}}$, at some communication round $t$. For the sake of clarity, we will drop specific indexing of communication rounds for now. The parameter $(i,j)$ in $\Delta w$ is computed as ${{\mu }_{i,j}}=\frac{1}{K}\sum\nolimits_{k=1}^{K}{\Delta w_{i,j}^{k}}$,where $\Delta w_{i,j}^{k}$ is the $(i,j)th$ parameter in the update of $\Delta {{w}^{k}}$, $k$ is the index of the client participating in the current round of federated learning, and $K$ is the number of clients participating in the current round of federated learning. We then define the update scales as the sum over all parameter variances in the updated matrix$\Delta w$:
\begin{equation}
    {U_s} = \frac{1}{{p \times q}}\sum\limits_{i = 0}^p {\sum\limits_{j = 0}^q {\mu _{i,j}^2} }
\end{equation}
It represents the extent of change in the global model during one round of communication.\\
\textbf{Definition 2:} The variance of parameters$\left( i,j \right)$throughout all $K$ clients is defined as:
\begin{equation}
    VAR\left[ {\Delta {w_{i,j}}} \right] = \frac{1}{K}{\sum\limits_{k = 0}^K {\left( {\Delta w_{i,j}^k - {\mu _{i,j}}} \right)} ^2}
\end{equation}
\textbf{Definition 3:} We the define ${{V}_{c}}$ as the sum over all parameter variances in the update matrix:
\begin{equation}
    {V_c} = \frac{1}{{q \times p}}\sum\limits_{i = 0}^q {\sum\limits_{j = 0}^p {VAR\left[ {\Delta {w_{i,j}}} \right]} }
\end{equation}
Geyer et al. \cite{r36} mentioned that federated learning can be divided into two stages: the label fitting stage and the data fitting stage. During the label fitting phase, client updates are more similar, so the sum over all parameter variances in the update matrix ${{V}_{c}}$ is relatively small, while the global model update scale ${{U}_{s}}$ is relatively large because there are significant updates to the randomly initialized weights. During the data fitting phase, ${{V}_{c}}$ gradually increases as each client optimizes towards its own dataset. At the same time, ${{U}_{s}}$ gradually decreases as it approaches the local optimum of the global model, with accuracy converging and contributions partially offsetting each other to some extent.\\
Li et al. [25] used this conclusion to propose a delayed fairness intervention method, utilizing the trend of the global model update scale ${{U}_{s}}$ to determine the intervention time. The specific method is as follows:
\begin{equation}
    \Delta U_s^t = \frac{1}{D}\sum\limits_{i = t - D}^t {U_s^i} - \frac{1}{D}\sum\limits_{i = t - D - 1}^{t - 1} {U_s^i} < \eta
\end{equation}
Where ${{U}_{s}}$ represents the global model update scale,$D$ represents the size of the sliding window,$t$ represents the current update round, and $\eta$ represents the threshold for determining whether to start the fairness intervention. It can be observed that FedGini requires computing the global model's update scale each round, which may introduce substantial computational overhead—especially in neural networks with a large number of parameters. This poses scalability challenges for training large language models under federated learning frameworks.
\begin{figure}[htbp]
  \centering
  \includegraphics[width=\linewidth]{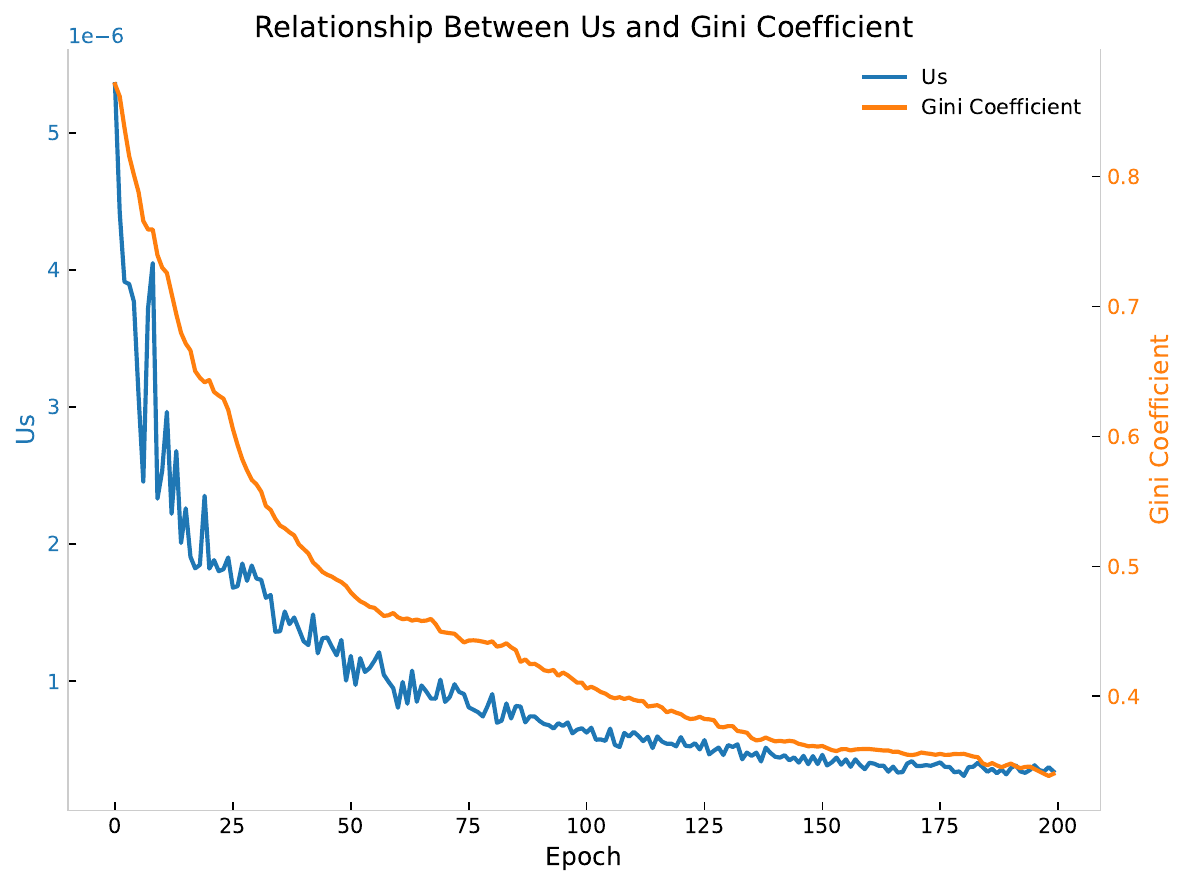}
  \caption{Relationship between Global Update Scale and Gini Coefficient.}
  \Description{}
  \label{fig:synthetic_0_0_relation}
\end{figure}

To mitigate this issue, we revisit the underlying relationship between ${{U}_{s}}$ and the Gini coefficient $G$. As illustrated in \textbf{Figure \ref{fig:synthetic_0_0_relation}} using the FedAvg algorithm on the \verb|Synthetic_0_0| dataset, the trajectories of ${{U}_{s}}$ and the Gini coefficient $G$ exhibit highly similar trends during training. Both metrics decrease substantially and almost synchronously during the early stages of training, suggesting that the Gini coefficient $G$ may serve as an efficient proxy for ${{U}_{s}}$ in determining the timing for fairness intervention.\\
Building on this insight, we propose a lightweight alternative by replacing ${{U}_{s}}$ with the Gini coefficient in the fairness trigger mechanism. The revised condition is given by:
\begin{equation}
    \Delta G = \frac{1}{D}\sum\limits_{i = t - 2D}^{t - D} {{G^i}}  - \frac{1}{D}\sum\limits_{i = t - D}^t {{G^i}} < \eta
\end{equation}
\subsection{Dynamic Aggregation Weight Adjustment Algorithm Based on Accuracy}
Under non-IID distributions, heterogeneous client data leads to discrepant local models. Standard aggregation favors high-quality data clients, marginalizing those with less representative data. To enhance fairness, we adapt aggregation weights based on the global model's performance on client validation sets. Our approach assigns higher weights to underperforming clients and lower weights to well-performing ones, with weight adjustments proportional to performance disparities. This mechanism ensures balanced representation of all clients' local models in the global model, particularly benefiting underrepresented data distributions. \\
The specific algorithm for dynamic weight adjustment is presented below:
\begin{equation}
    weigh{t_i} = 1 - {a_i}
\end{equation}
\begin{equation}
    weigh{t_i} = \frac{{{\rm{weigh}}{{\rm{t}}_i}}}{{\sum\limits_{i = 1}^n {{\rm{weigh}}{{\rm{t}}_i}} }} \times \lambda
\end{equation}
\begin{equation}
    {\exp _i} = {e^{weigh{t_i}}}
\end{equation}
\begin{equation}
    weigh{t_i} = \frac{{{{\exp }_i}}}{{\sum\limits_{i = 1}^n {{{\exp }_i}} }}
\end{equation}
Where $weigh{{t}_{i}}$ is the aggregation weight of the $ith$ device, and ${{a}_{i}}$ is the validation accuracy of the $ith$ device. ${{\exp }_{i}}$ represents the $weigh{{t}_{i}}$ power of $e$. Equations (10) and (11) are designed to decrease the proportion of aggregation weights for better-performing clients and increase the proportion for worse-performing clients. A $softmax$ normalization is then used to magnify the weights of lower-accuracy devices, allowing the global model to learn more from these devices during aggregation, thereby encouraging the global model to learn more from underperforming clients and improving overall fairness. The hyperparameter $\lambda $ controls the strength of fairness intervention: a larger $\lambda $ results in more emphasis on fairness, with values $\lambda >1$ typically used in practice. 
\begin{figure}[htbp]
  \centering
  \includegraphics[width=\linewidth]{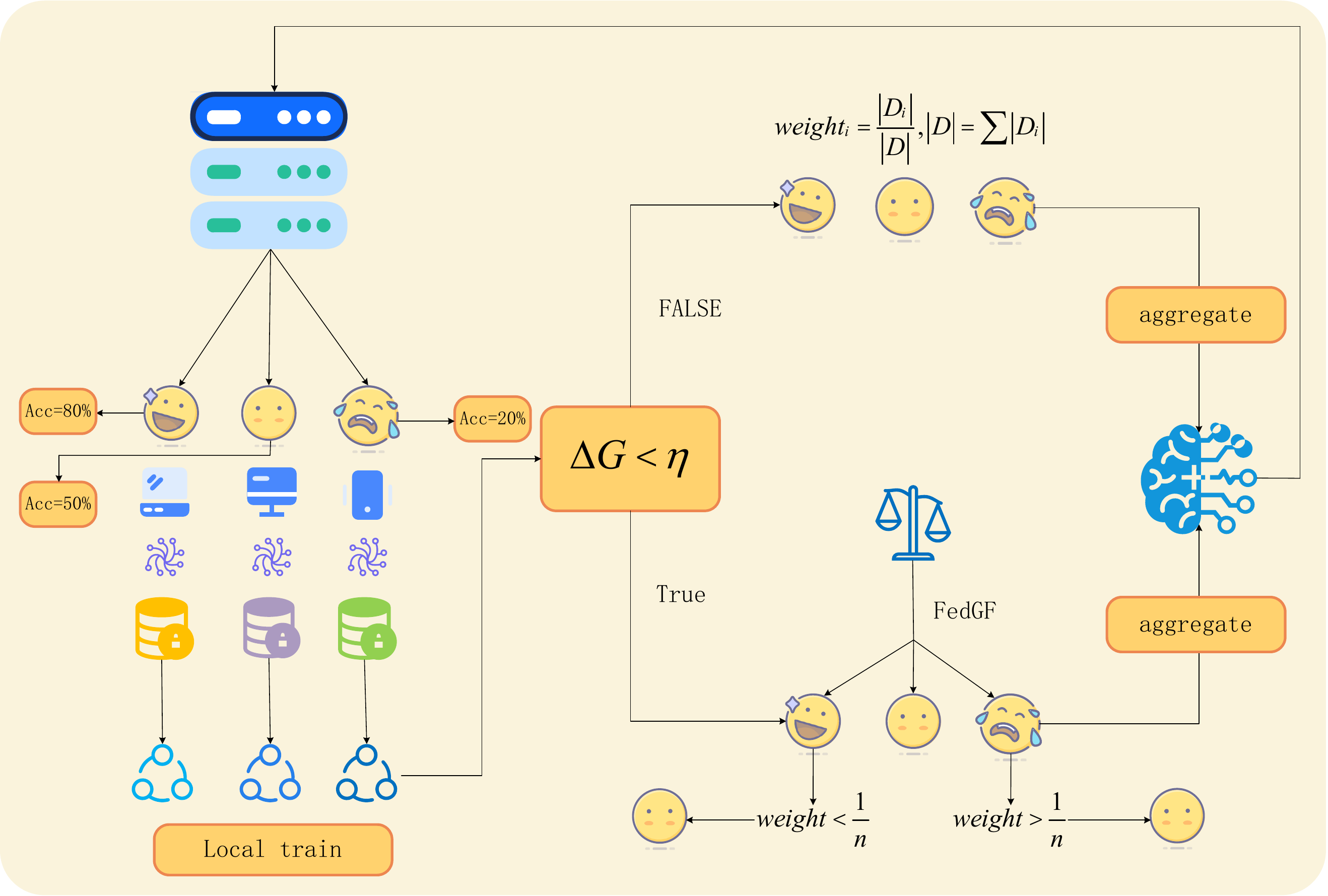}
  \caption{Overview of the FedGA Algorithm Workflow.}
  \Description{}
  \label{fig:FedGA_flowchart}
\end{figure}

To provide an intuitive understanding of the proposed method, \textbf{Figure \ref{fig:FedGA_flowchart}} illustrates the overall workflow of FedGA. The algorithm first evaluates whether fairness intervention should be applied by monitoring the change in the Gini coefficient across communication rounds. If the change falls below a predefined threshold, fairness-aware aggregation is triggered, assigning higher weights to underperforming clients to improve fairness in the global model. Otherwise, standard aggregation is performed. Algorithm 1 presents the pseudocode of FedGA.
\begin{algorithm}
\caption{\textbf{Gini Coefficient-aware Fair Federated Learning (FedGA)}}
\begin{flushleft}
\textbf{Input:} Number of communication rounds $T$, number of local iterations $E$, initial aggregation weight $p$ \\
\textbf{Output:} Optimal global model $W_{\text{op}}$
\end{flushleft}
\begin{algorithmic}[1]
\State $t \gets 0$
\While{$t \leq T - 1$}
    \State \textbf{Client}($W_s$):
    \State \quad Evaluate the global model on validation set $a_k \gets W_s(D_k)$
    \For{$e = 1$ to $E$}
        \State $W^k_{t+1} \gets W_s - \eta \nabla F(W_s)$
    \EndFor
    \State Return $W^k_{t+1}$ and $a_k$
    \State \textbf{Server:}
        \State Randomly select $m$ clients from $M$: $S_t \subset M$
        \For{each client $k \in S_t$ in parallel}
            \State $W^k_{t+1} \gets \text{Client}(W_s)$
        \EndFor
        \State Server updates accuracy list $Acc = [a_1, ..., a_m]$
        \State Compute Gini coefficient:
        \[
        G = \frac{\sum_{i=1}^n \sum_{j=1}^n |x_i - x_j|}{2(n - 1)\sum_{j=1}^n x_j}
        \]
        \State Compute $\Delta G = \frac{1}{D}\sum\limits_{i = t - 2D}^{t - D} {{G^i}} - \frac{1}{D}\sum\limits_{i = t - D}^t {{G^i}}$
        \If{$\Delta G < \eta$}
            \State $weight_i = 1 - a_i$
            \State $weight_i \gets \frac{weight_i}{\sum_{j=1}^{n} weight_j} \times \lambda$
            \State $exp_i \gets e^{weight_i}$
            \State $weight_i \gets \frac{exp_i}{\sum_{j=1}^{n} exp_j}$
        \Else
            \State $W^{t+1} \gets \sum_{k \in S_t} w_k^a \cdot W^{t+1}_k,\quad w_k^a = \frac{n_k}{n}$
        \EndIf
\EndWhile
\end{algorithmic}
\end{algorithm}


\section{Theoretical Analysis}
In this section, we conduct a series of theoretical analyses. Section 4.1 analyzes the communication overhead of our method and compares it with that of FedAvg. Section 4.2 evaluates the time complexity of FedGA in identifying the optimal intervention timing, and compares it with FedGini. In Section 4.3, we prove that FedGA consistently assigns an aggregation weight greater than $\frac{1}{n}$ to the worst-performing client and less than $\frac{1}{n}$ to the best-performing client. Section 4.4 explores the relationship between the Gini coefficient and the mean of the total performance disparity among clients.
\subsection{Communication Overhead Analysis}
Compared to the Federated Averaging algorithm (FedAvg), FedGA introduces only a minimal communication overhead by requiring each client to upload the accuracy of the global model on its local validation set. Assuming this accuracy is represented using single-precision floating-point format, each value occupies $4B$ of memory. Since this information is only transmitted from the client to the server and does not need to be returned, the additional communication overhead is limited to $4B$ per client.\\
In contrast, the communication overheads for the AlexNet and ResNet-18 models used in our experiments are $2\times 49.5MB=99MB$ and $2\times 44.6MB=89.2MB$, respectively. Therefore, we conclude that FedGA introduces negligible communication overhead and does not impose a significant burden on federated learning systems.
\subsection{Algorithm Complexity Analysis}
To compare the time complexity between the delayed fair intervention method proposed by FedGA and that by FedGini. We provide the following analysis:\\
FedGini requires the calculation of the following two formulas:
\begin{equation}
    {\mu _{i,j}} = \frac{1}{K}\sum\nolimits_{k = 1}^K {\Delta w_{i,j}^k}
    \label{eq:mu}
\end{equation}
\begin{equation}
    {U_s} = \frac{1}{{p \times q}}\sum\limits_{i = 0}^p {\sum\limits_{j = 0}^q {\mu _{i,j}^2} }
    \label{eq:Us}
\end{equation}
Formula \ref{eq:mu} includes a summation operation, summing from $k=1$ to $k=k$, a total of $k$ times. For each combination of $i$ and $j$, the summation operation runs $k$ times. The time required for each summation operation is a constant time operation (i.e., calculating $\Delta w_{i,j}^{k}$ and adding it to the sum). Therefore, the time complexity of the entire summation operation can be expressed as $O\left( K \right)$.\\
Formula \ref{eq:Us} includes two nested summation operations. The outer summation runs from $i=0$ to $i=p$, a total of $p+1$ times. The inner summation runs from $j=0$ to $j=q$, a total of $q+1$ times. For each value of $i$, the inner summation operation runs $q+1$ times. Thus, the total number of summation operations is $\left( p+1 \right)\times \left( q+1 \right)$ times. Therefore, the time complexity of the entire summation operation can be expressed as $O\left( \left( q+1 \right)\times \left( p+1 \right) \right)$. Since constant factors can be ignored in Big O notation, the time complexity simplifies to $O\left( q\times p \right)$.\\
However, in formula \ref{eq:Us}, each calculation of ${{\mu }_{i,j}}$ involves formula \ref{eq:mu}. Since the calculation of ${{\mu }_{i,j}}$ requires $O\left( K \right)$ time, the time for calculating each ${{\mu }_{i,j}}$ in formula \ref{eq:Us} will also be $O\left( K \right)$. Therefore, the time complexity of computing the entire formula (15) is $O\left( q\times p \right)\times O\left( k \right)=O\left( q\times p\times k \right)$. For ease of comparison with FedGA, the time complexity of FedGini is expressed as $O\left( q\times p\times n \right)$.\\
FedGA requires the calculation of the Gini coefficient, and the time complexity of this calculation is as follows:\\
The numerator includes two nested summation operations: $\sum\limits_{i=1}^{n}{\sum\limits_{j=1}^{n}{\left| {{x}_{i}}-{{x}_{j}} \right|}}$. The outer summation runs from $i=1$ to $i=n$, a total of $n$ times, and the inner summation runs from $j=1$ to $j=n$, also a total of $n$ times. Therefore, the total number of summation operations is $n\times n={{n}^{2}}$ times. Calculating $\left| {{x}_{i}}-{{x}_{j}} \right|$ is a constant time operation (assuming the absolute value operation is $O\left( 1 \right)$). Thus, the time complexity of the numerator is $O\left( {{n}^{2}} \right)$.The denominator includes two operations: $2\left( \text{n-}1 \right)\sum\limits_{j=1}^{n}{{{x}_{j}}}$. Calculating $\sum\limits_{j=1}^{n}{{{x}_{j}}}$ requires summing $n$ elements, with a time complexity of $O\left( n \right)$. Multiplying by $2\left( \text{n-}1 \right)$ is a constant time operation, $O\left( 1 \right)$. Thus, the time complexity of the denominator is $O\left( n \right)$. \\
Since the time complexity of the denominator $O\left( n \right)$ is lower than that of the numerator $O\left( {{n}^{2}} \right)$, the overall time complexity is determined by the numerator. Therefore, the total time complexity is $O\left( {{n}^{2}} \right)$.\\
In this context, $k$ and $n$ in the algorithm complexity represent the number of clients participating in federated learning training. When the number of neural network parameters $q\times p$ is greater than $n$,the computational overhead of FedGA computational overhead is lower than that of FedGini \cite{r25}.
\subsection{Analysis of Aggregation Weights}
Let the set of clients participating in the current round of federated learning training be $N$. For client $z$, after the computation of the dynamic aggregation adjustment algorithm, the weight is:
\begin{equation}
    Weigh{t_z} = \frac{{{e^{\frac{{{x_z}}}{{\sum\limits_{j = 1}^n {{x_j}} }} \times \lambda }}}}{{\sum\limits_{i = 1}^n {{e^{\frac{{{x_i}}}{{\sum\limits_{j = 1}^n {{x_j}} }} \times \lambda }}} }}
\end{equation}
Where $n$ is the number of clients, and ${{x}_{z}}=1-{{\alpha }_{z}}$. \\
Dividing both the numerator and the denominator by the numerator, we get:
\begin{equation}
    Weigh{t_z} = \frac{1}{{1 + \frac{{\sum\limits_{i \ne z}^n {{e^{\frac{{{x_i}}}{{\sum\limits_{j = 1}^n {{x_j}} }} \times \lambda }}} }}{{{e^{\frac{{{x_z}}}{{\sum\limits_{j = 1}^n {{x_j}} }} \times \lambda }}}}}}
\end{equation}
According to the laws of exponents, we can derive:
\begin{equation}
    Weigh{t_z} = \frac{1}{{1 + \sum\limits_{i \ne z}^n {{e^{\frac{{\lambda  \times \left( {{x_i} - {x_z}} \right)}}{{\sum\limits_{j = 1}^n {{x_j}} }}}}} }}
\end{equation}
Assuming that client $z$ is the best-performing client, then ${{x}_{z}}$ is the smallest among all clients, and $\lambda >0$. For $\forall i\in N\backslash \left\{ z \right\}$, we have:
\begin{equation}
    \lambda  \times \left( {{x_i} - {x_z}} \right) > 0
\end{equation}
Therefore:
\begin{equation}
    \frac{{\lambda  \times \left( {{x_i} - {x_z}} \right)}}{{\sum\limits_{j = 1}^n {{x_j}} }} > 0
\end{equation}
According to the properties of the exponential function with base $e$, we obtain:
\begin{equation}
    {e^{\frac{{\lambda  \times \left( {{x_i} - {x_z}} \right)}}{{\sum\limits_{j = 1}^n {{x_j}} }}}} > 1
\end{equation}
Therefore:
\begin{equation}
    1 + \sum\limits_{i \ne z}^n {{e^{\frac{{\lambda  \times \left( {{x_i} - {x_z}} \right)}}{{\sum\limits_{j = 1}^n {{x_j}} }}}}}  > n
\end{equation}
Therefore:
\begin{equation}
    Weigh{t_z} = \frac{1}{{1 + \sum\limits_{i \ne z}^n {{e^{\frac{{\lambda  \times \left( {{x_i} - {x_z}} \right)}}{{\sum\limits_{j = 1}^n {{x_j}} }}}}} }} < \frac{1}{n}
\end{equation}
Assuming that client $z$ is the worst-performing client, then ${{x}_{z}}$ is the largest among all clients, and $\lambda >0$. For $\forall i\in N\backslash \left\{ z \right\}$, we have:
\begin{equation}
    \lambda  \times \left( {{x_i} - {x_z}} \right) < 0
\end{equation}
Therefore:
\begin{equation}
    \frac{{\lambda  \times \left( {{x_i} - {x_z}} \right)}}{{\sum\limits_{j = 1}^n {{x_j}} }} < 0
\end{equation}
According to the properties of the exponential function with base $e$, we obtain:
\begin{equation}
    {e^{\frac{{\lambda  \times \left( {{x_i} - {x_z}} \right)}}{{\sum\limits_{j = 1}^n {{x_j}} }}}} < 1
\end{equation}
Therefore:
\begin{equation}
    1 + \sum\limits_{i \ne z}^n {{e^{\frac{{\lambda  \times \left( {{x_i} - {x_z}} \right)}}{{\sum\limits_{j = 1}^n {{x_j}} }}}}}  > n
\end{equation}
Therefore:
\begin{equation}
    Weigh{t_z} = \frac{1}{{1 + \sum\limits_{i \ne z}^n {{e^{\frac{{\lambda  \times \left( {{x_i} - {x_z}} \right)}}{{\sum\limits_{j = 1}^n {{x_j}} }}}}} }} > \frac{1}{n}
\end{equation}

\subsection{Proof of the Relationship Between the Gini Coefficient and the Mean Accuracy Disparity Among Clients in Federated Learning}
\textbf{Definition 4}: The average sum of accuracy differences among clients, denoted as $AvgDiff=\frac{\sum\limits_{i\ne j}{\left| {{x}_{i}}-{{x}_{j}} \right|}}{n(n-1)}$, is introduced to characterize the consistency of client performance and the generalization ability of the global model under heterogeneous data distributions.\\
Since:
\begin{equation}
    \sum\limits_{i \ne j} {\left| {{x_i} - {x_j}} \right|}  = \sum\nolimits_{i = 1}^n {\sum\nolimits_{j = 1}^n {\left| {{x_i} - {x_j}} \right|} }  - \sum\limits_i {\left| {{x_i} - {x_i}} \right|}  = \sum\nolimits_{i = 1}^n {\sum\nolimits_{j = 1}^n {\left| {{x_i} - {x_j}} \right|} }
\end{equation}
According to Equation \ref{eq:gini}, the Gini coefficient can be expressed as:
\begin{equation}
    G = \frac{{\sum\nolimits_{i = 1}^n {\sum\nolimits_{j = 1}^n {\left| {{x_i} - {x_j}} \right|} } }}{{2(n - 1)\sum\nolimits_{i = 1}^n {{x_i}} }} = \frac{{\sum\nolimits_{i = 1}^n {\sum\nolimits_{j = 1}^n {\left| {{x_i} - {x_j}} \right|} } }}{{2n(n - 1)\mu }} = \frac{{\sum\limits_{i \ne j} {\left| {{x_i} - {x_j}} \right|} }}{{2n(n - 1)\mu }}
\end{equation}
where $\mu  = \frac{{\sum\nolimits_{i = 1}^n {{x_i}} }}{n}$ denotes the mean accuracy across all clients.\\
Therefore, $AvgDiff\left( {\mu ,G} \right) = 2\mu G$.
Taking the total differential of $AvgDiff\left( {\mu ,G} \right)$, we have:
\begin{equation}
    dAvgDiff = \frac{{\partial AvgDiff}}{{\partial \mu }}\;d\mu  + \frac{{\partial AvgDiff}}{{\partial G}}\;dG
\end{equation}
By computing the partial derivatives, it follows that:
\begin{equation}
    \frac{{\partial AvgDiff}}{{\partial \mu }} = 2G, \frac{{\partial AvgDiff}}{{\partial G}} = 2\mu
\end{equation}
Thus, $dAvgDiff = 2(\mu dG + Gd\mu )$.
By performing a first-order Taylor expansion at the point $\left( {\mu ,G} \right)$, we obtain:
\begin{equation}
    AvgDiff(\mu  + \Delta \mu ,G + \Delta G) \approx AvgDiff(\mu ,G) + \frac{{\partial AvgDiff}}{{\partial \mu }}\Delta \mu  + \frac{{\partial AvgDiff}}{{\partial G}}\Delta G
\end{equation}
Therefore,$\Delta AvgDiff \approx 2G\Delta \mu  + 2\mu \Delta G = 2(\mu \Delta G + G\Delta \mu )$.\\
The approximation error is of the order $O\left( {\Delta \mu \Delta G,\Delta {\mu ^2},\Delta {G^2}} \right)$, and the first-order expansion is valid when higher-order terms are negligible, such as in the late stage of training where fluctuations are small.\\
Assuming that in the late stage of training the mean accuracy $\mu$ is 70\% and the Gini coefficient $G$ is 0.1, when $\Delta G$ decreases by 0.01 and $\Delta \mu $ increases by 0.01, we have:
\begin{equation}
    \Delta AvgDiff \approx 2 \times \left( {0.7 \times  - 0.01 + 0.1 \times 0.01} \right) =  - 0.012
\end{equation}
This indicates that the average pairwise accuracy difference among clients is reduced by approximately 1.2\%.

\section{Experiments}

In this section, we present the empirical results to demonstrate the effectiveness of the proposed FedGA algorithm. Section 5.1 details the datasets and experimental setup. Section 5.2 provides an analysis of the main experimental results. In Section 5.3, we investigate the impact of hyperparameters. Section 5.4 presents the results of the ablation studies. Finally, Section 5.5 compares the computational efficiency of FedGA with that of FedGini.\footnote{The source code will be released soon.}
\subsection{Datasets}
We use two real-world datasets and one synthetic dataset: Office-Caltech-10 \cite{r32}, CIFAR-10 \cite{r31}, and synthetic \cite{r5}. The Office-Caltech-10 dataset simulates a feature heterogeneous scenario, while the CIFAR-10 dataset simulates a label heterogeneous scenario.\\
\textbf{Experimental Details}: We conducted experiments on the Office10 dataset using two different network architectures: AlexNet and ResNet18, referred to as office10\_alexnet and office10\_resnet18, respectively. For the experiments with AlexNet, we set the learning rate to 0.01 and the batch size to 32, while for ResNet18, the learning rate was set to 0.1 with a batch size of 64. In both cases, the Stochastic Gradient Descent (SGD) optimizer was employed, with one local epoch per round and a total of 400 communication rounds. For the CIFAR-10 dataset, we adopted the ResNet18 architecture \cite{r34}, using the SGD optimizer with a learning rate of 0.1, a batch size of 64, one local epoch, and a total of 600 communication rounds. The data were partitioned using a Dirichlet distribution with a concentration parameter of 0.1, referred to as CIFAR-01. On the synthetic dataset, We employ a simple linear classification model consisting of a single fully connected layer that maps the input feature vector to the output class logits. The learning objective was to optimize parameters $W$ and $b$, using the SGD optimizer with a learning rate of 0.01, a batch size of 32, one local epoch, and 200 communication rounds. For all experiments, we performed five independent runs with different random seeds and reported the mean and standard deviation of the results.\\
\textbf{Baselines:} We compare our method against the following representative baselines: FedAvg \cite{r2}, AFL \cite{r29}, FedProx \cite{r5}, q-FedAvg \cite{r3}, FedFa \cite{r9}, Fedmgda+ \cite{r30}, FedFV \cite{r26}, and FedGini \cite{r25}. To ensure a fair comparison, we adopted the hyperparameter configurations summarized in \textbf{Table \ref{tab:hyperparameters}} to validate all methods, and reported the best performance achieved by each. 
\begin{table}[htbp]
  \caption{Hyperparameters of Baseline Methods}
  \label{tab:hyperparameters}
  \begin{tabular}{cc}
    \toprule
    Method & Parameters\\
    \midrule
    AFL & ${\eta _\lambda } \in \left\{ {0.01,0.1,0.5} \right\}$\\
    q-FedAvg & $q \in \left\{ {0.1,0.2,1.0,2.0,5.0} \right\}$\\
    FedFa & $\beta  \in \left\{ {0.0,0.5,1.0} \right\}$\\
    Fedmgda+ & $\varepsilon  \in \left\{ {0.01,0.05,0.1,0.5,1.0} \right\}$\\
    FedFV & $\alpha  \in \left\{ {0.1,0.3,0.5,1.0} \right\},\tau  \in \left\{ {0,1,3,10} \right\}$\\
    FedGini & $\varepsilon  \in \left\{ {0,0.5,1} \right\}$\\
  \bottomrule
\end{tabular}
\end{table}
\subsection{Experimental Results Analysis}
In this section, we present experimental results on three datasets and conduct a detailed analysis. For clarity and fairness, all figures and tables report the performance of each baseline under its best-performing hyperparameter settings.
\begin{figure}[htbp]
  \centering
  \includegraphics[width=\linewidth]{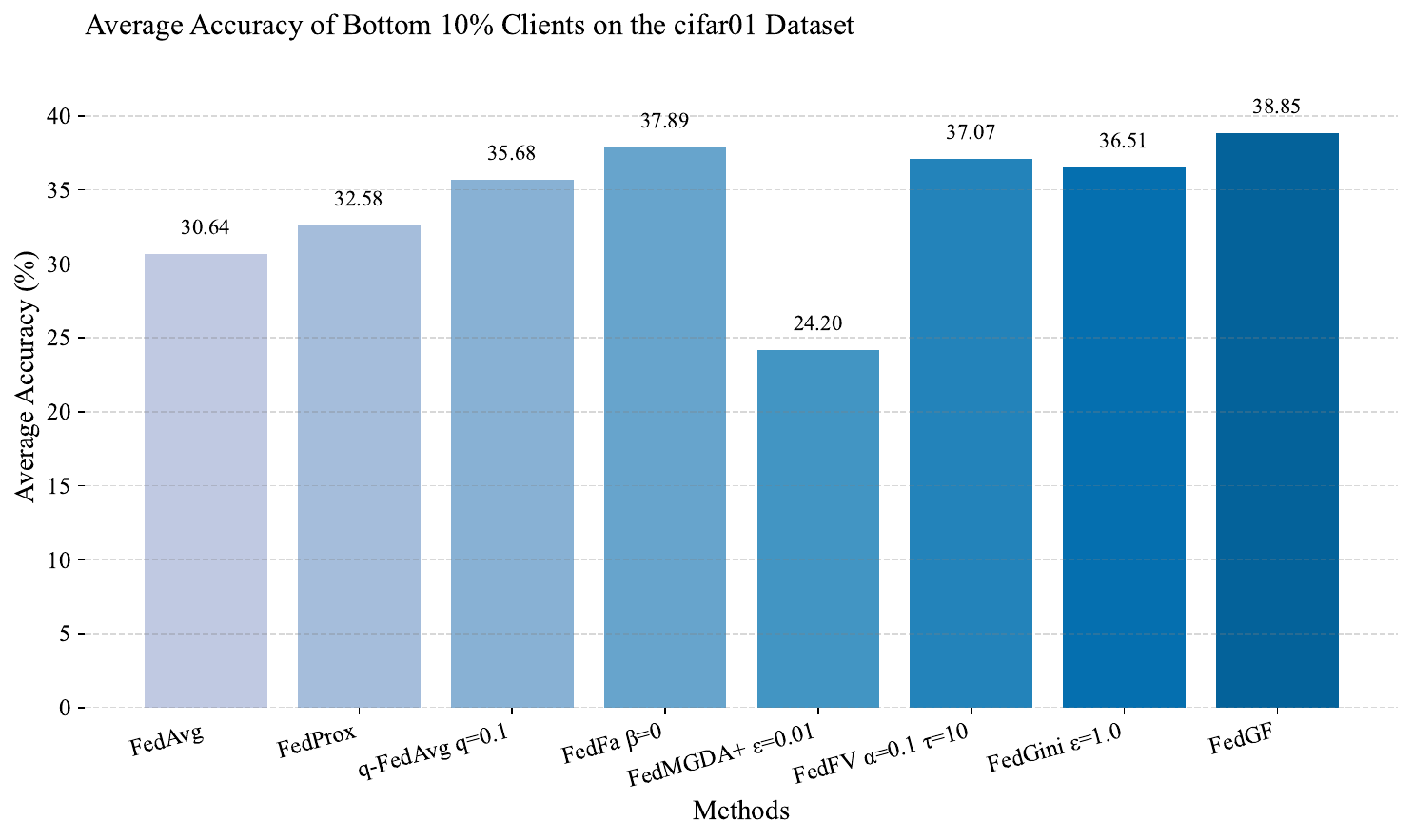}
  \caption{Average Test Accuracy of the Bottom 10\% Clients on the CIFAR-01 Dataset.}
  \Description{}
  \label{fig:cifar01_bottom_10}
\end{figure}

\textbf{Figure \ref{fig:cifar01_bottom_10}} presents the average test accuracy of the bottom 10\% of clients across various federated learning methods on the cifar01 dataset. This metric highlights how well different algorithms serve the most disadvantaged clients. FedGA achieves the highest bottom-10\% accuracy (38.85±3.41\%), outperforming all baselines. This indicates that FedGA provides support for clients in the worst-case regime, ensuring that even clients with adverse data conditions receive a model that performs reliably. Compared to FedAvg (30.64±2.22\%) and FedProx (32.58±3.62\%),  FedGA improves tail client accuracy by approximately 6-8 percentage points, indicating a meaningful reduction in client-level disparity.
\begin{figure}[htbp]
  \centering
  \includegraphics[width=\linewidth]{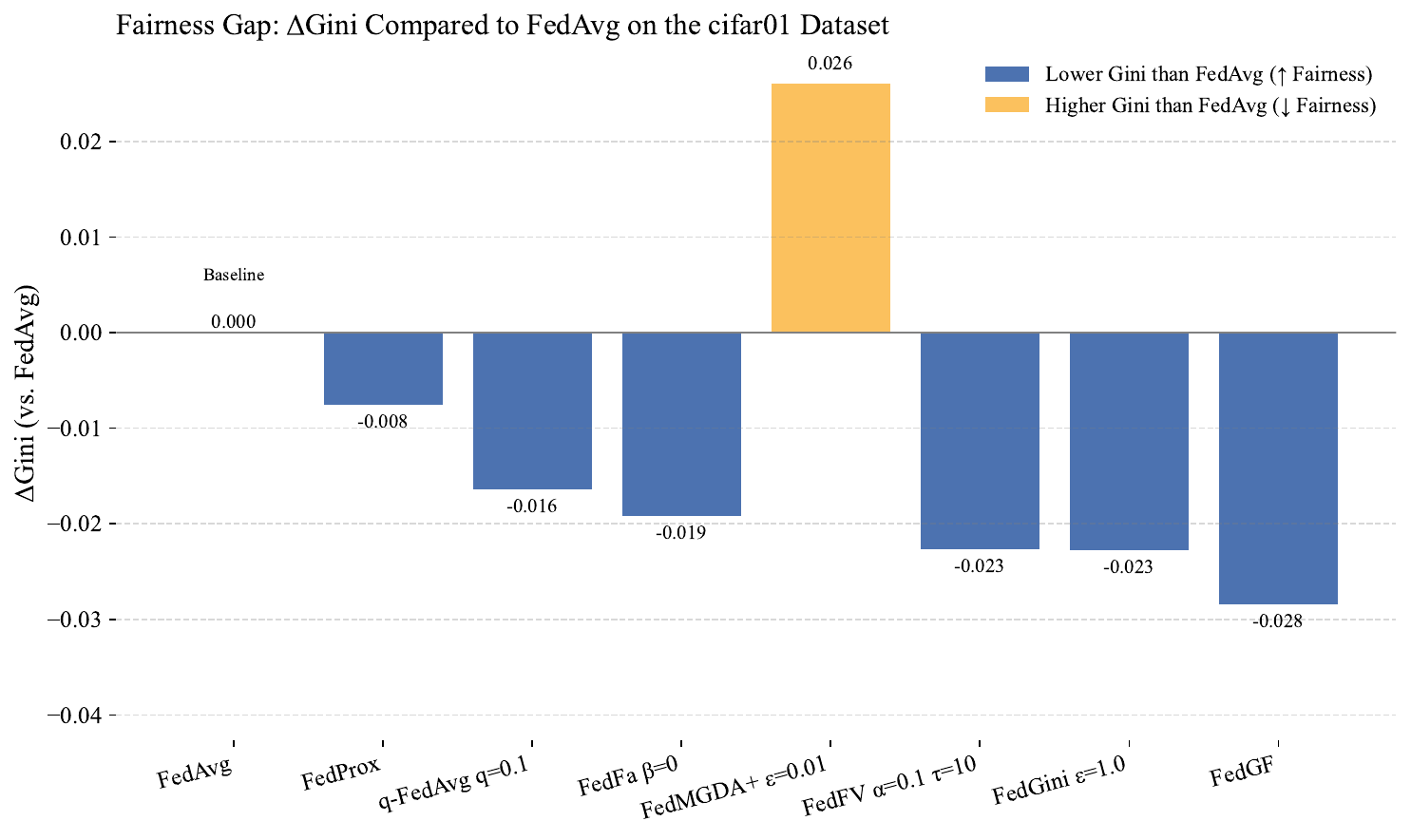}
  \caption{Fairness Gap Measured by $\Delta$Gini Relative to FedAvg on the CIFAR-01 Dataset.}
  \Description{}
  \label{fig:cifar01_delta}
\end{figure}

\textbf{Figure \ref{fig:cifar01_delta}} shows $\Delta$Gini values for federated learning algorithms on the FedAvg baseline. $\Delta$Gini quantifies the Gini coefficient difference between methods and FedAvg; negative values indicate improved fairness. FedGA achieves the largest improvement ($\Delta$Gini = -0.028), representing substantial fairness enhancement while confirming its effectiveness in reducing performance gaps across heterogeneous federated settings. Other fairness-enhancing methods show smaller improvements: FedFV and FedGini achieve $\Delta$Gini = -0.023, indicating moderate inequality reductions. Meanwhile, FedFa demonstrates minimal improvement ($\Delta$Gini = -0.019) and q-FedAvg ($\Delta$Gini = -0.016) shows slight fairness degradation, indicating less improvement relative to FedAvg.
\begin{table}[htbp]
  \caption{Comparison of Client Accuracy standard deviation Across Federated Learning Methods on the Cifar-01 dataset}
  \label{tab:cifar01_std}
  \begin{tabular}{lc}
    \toprule
    Method & Std\\
    \midrule
    FedAvg & 14.49±0.80\\
    FedProx & 13.65±1.65\\
    q-FedAvg~|~$q = 0.1$ & 12.89±0.96\\
    FedFa~|~$\beta  = 0$ & 13.24±1.71\\
    FedMgda+~|~$\varepsilon  = 0.01$ & 13.78±1.03\\
    FedFV~|~$\alpha  = 0.1,\tau  = 10$ & 12.17±1.48\\
    FedGini~|~$\varepsilon  = 1.0$ & 12.18±0.79\\
    \textbf{FedGA} & \textbf{11.54±1.48}\\
  \bottomrule
\end{tabular}
\end{table}

\textbf{Table \ref{tab:cifar01_std}} reports the standard deviation (Std) of client accuracies across various federated learning algorithms on the Cifar01 dataset. With a standard deviation of 11.54 ± 1.48, FedGA outperforms all baselines in minimizing inter-client performance variability. Algorithms such as FedFV (12.17 ± 1.48), FedFa (13.24 ± 1.71), and q-FedAvg (12.89 ± 0.96) also reduce variability relative to conventional methods but still fall short of FedGA’s level of uniformity.
\begin{figure}[htbp]
  \centering
  \includegraphics[width=\linewidth]{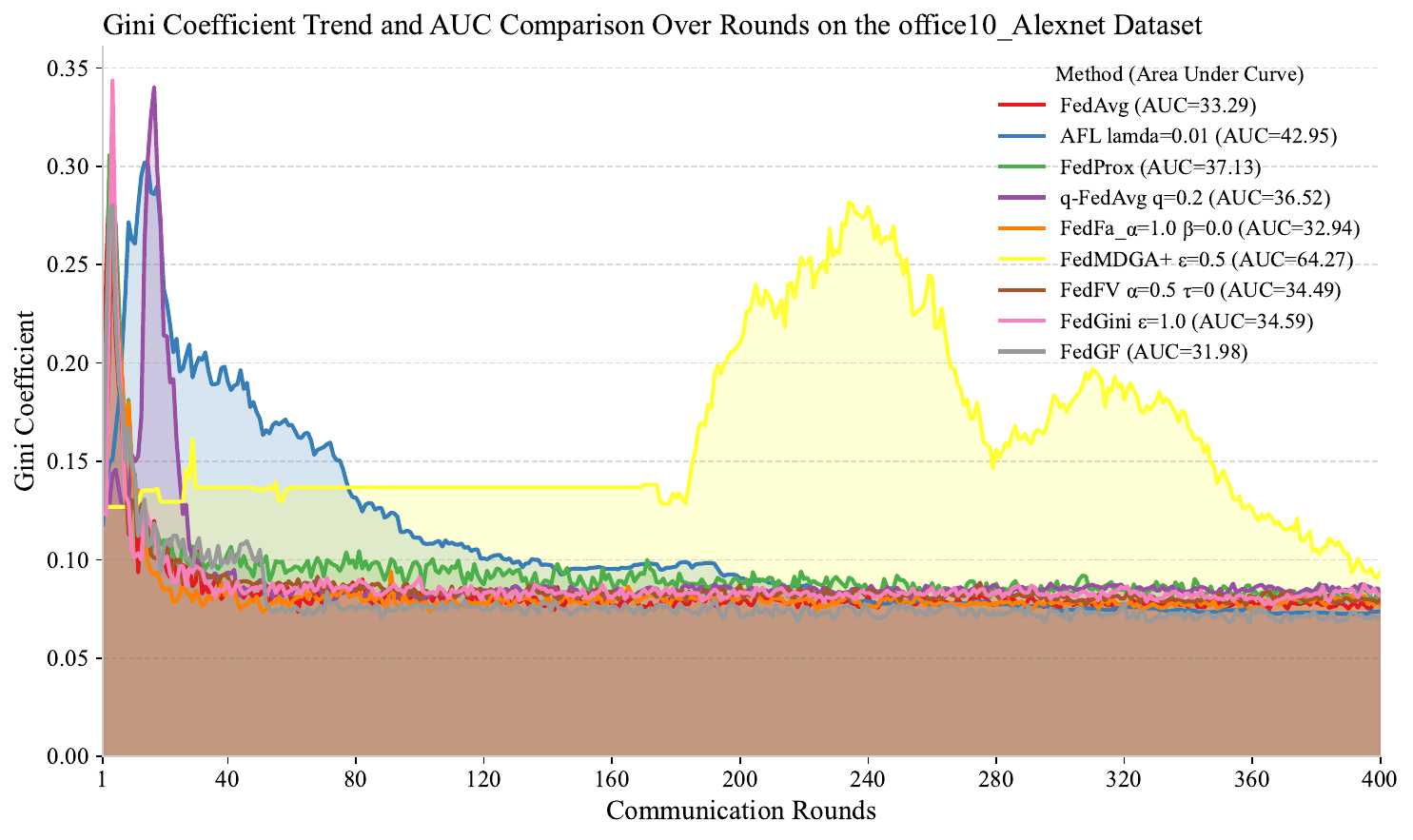}
  \caption{Smoothed Gini Coefficient Trajectories and AUC-Based Fairness Comparison on the Office10\_Alexnet Dataset.}
  \Description{}
  \label{fig:office10_auc}
\end{figure}

\textbf{Figure \ref{fig:office10_auc}} shows smoothed Gini coefficient trajectories over communication rounds with corresponding Area Under Curve (AUC) values for federated learning algorithms on the office10 dataset. FedGA achieves the lowest AUC(31.98), indicating superior cumulative fairness compared to FedAvg(33.29), FedProx(37.13), FedGini(34.59), and FedFV(34.49). FedGA demonstrates notable Gini coefficient reduction during training, consistently reaching low levels($\approx$0.1) after 50 rounds, suggesting effective fairness enhancement and reduced client performance disparities. These results highlight FedGA's potential for addressing feature heterogeneity challenges in federated learning.
\begin{figure}[htbp]
  \centering
  \includegraphics[width=\linewidth]{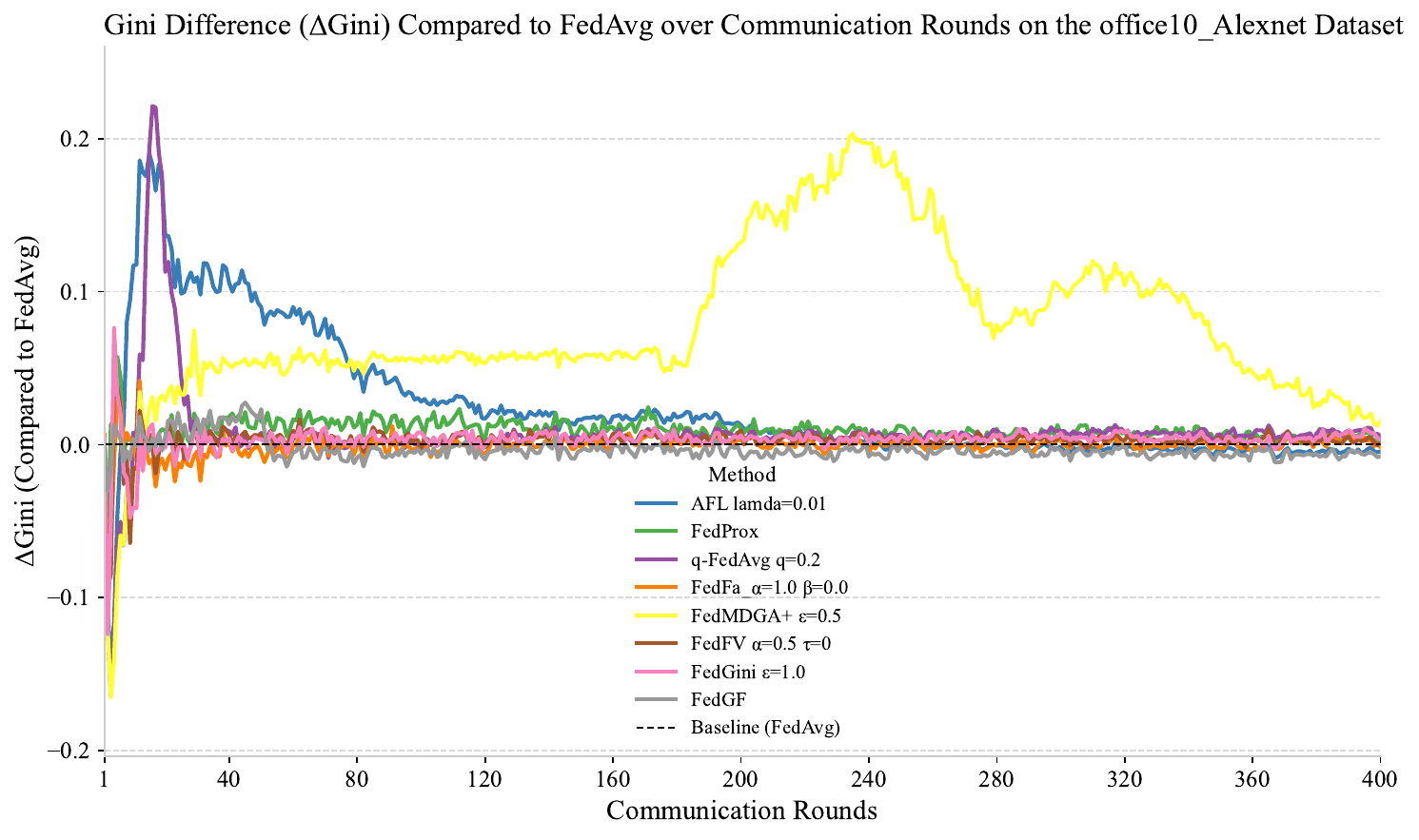}
  \caption{Smoothed Fairness Gap Trajectories ($\Delta$Gini) Relative to FedAvg on the Office10\_Alexnet Dataset.}
  \Description{}
  \label{fig:office10_delta_gini_curve}
\end{figure}

\textbf{Figure \ref{fig:office10_delta_gini_curve}} illustrates the smoothed $\Delta$Gini trajectories for various federated learning algorithms compared to FedAvg on the office10\_alexnet dataset over 400 communication rounds. $\Delta$Gini measures the Gini coefficient difference between each method and FedAvg, with negative values indicating reduced client performance inequality. FedGA consistently maintains negative $\Delta$Gini values throughout most training rounds, demonstrating superior fairness performance. In contrast, other fairness-oriented methods (FedFV and FedGini) exhibit predominantly positive $\Delta$Gini values, suggesting ineffective fairness protection, potentially due to their lack of optimization for full client participation scenarios.
\begin{figure}[htbp]
  \centering
  \includegraphics[width=\linewidth]{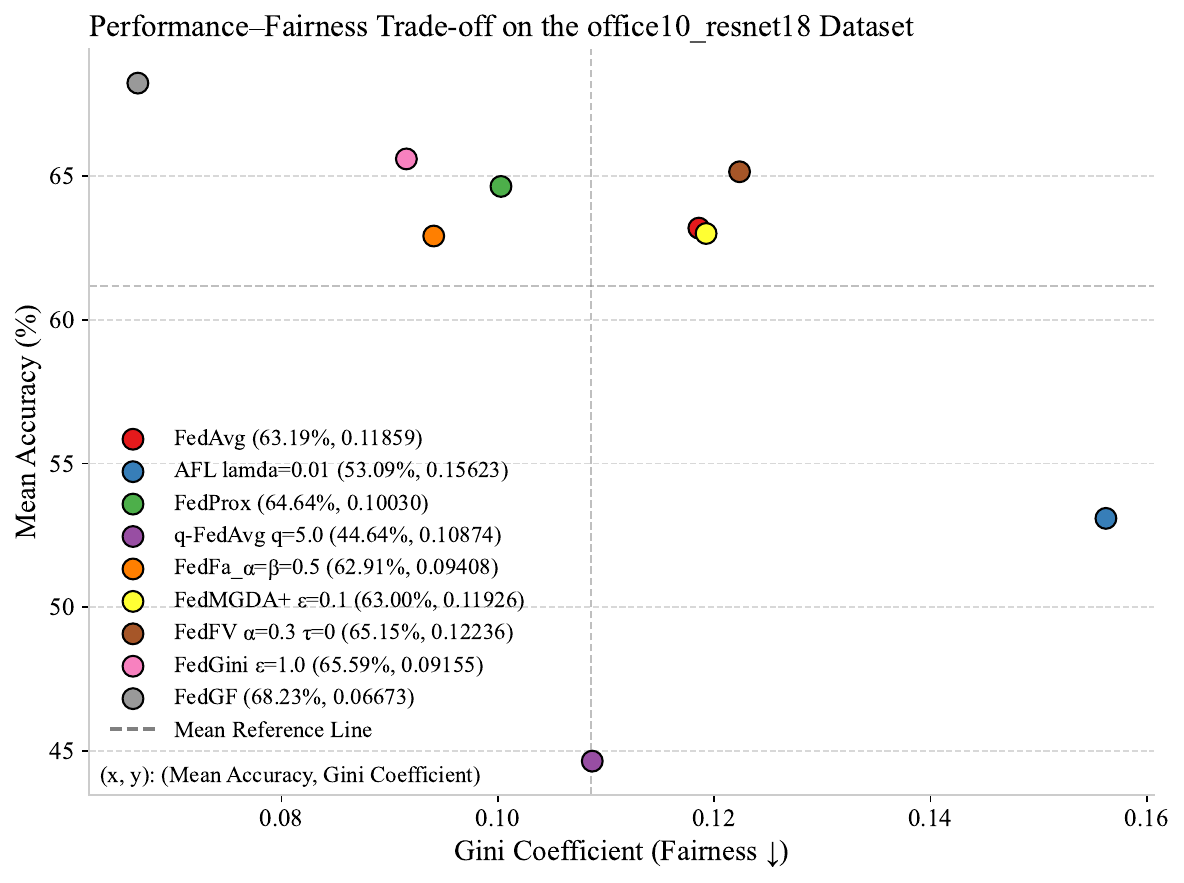}
  \caption{Performance–Fairness Trade-off of Federated Learning Algorithms on the Office10 Dataset (ResNet18 network).}
  \Description{}
  \label{fig:office10_fairness_performance}
\end{figure}

\textbf{Figure \ref{fig:office10_fairness_performance}}depicts the fairness–performance trade-off for different federated learning algorithms on the Office10 dataset using ResNet18. Each point represents a method's average test accuracy versus Gini coefficient. FedGA achieves the optimal balance with the highest accuracy (68.23$\pm$0.28\%) and lowest Gini coefficient (0.06673$\pm$0.01113), effectively promoting both global utility and fairness. Fairness-aware baselines (FedGini: 65.59$\pm$1.78\%, 0.09155$\pm$0.02196; FedFa: 62.91$\pm$4.37\%, 0.09408$\pm$0.03066) show improved fairness compared to FedAvg but at the cost of reduced accuracy. These results demonstrate FedGA's ability to mitigate client-level performance disparities without sacrificing global performance, while other fairness-oriented methods face inherent trade-offs between these objectives.
\begin{figure}[htbp]
  \centering
  \includegraphics[width=\linewidth]{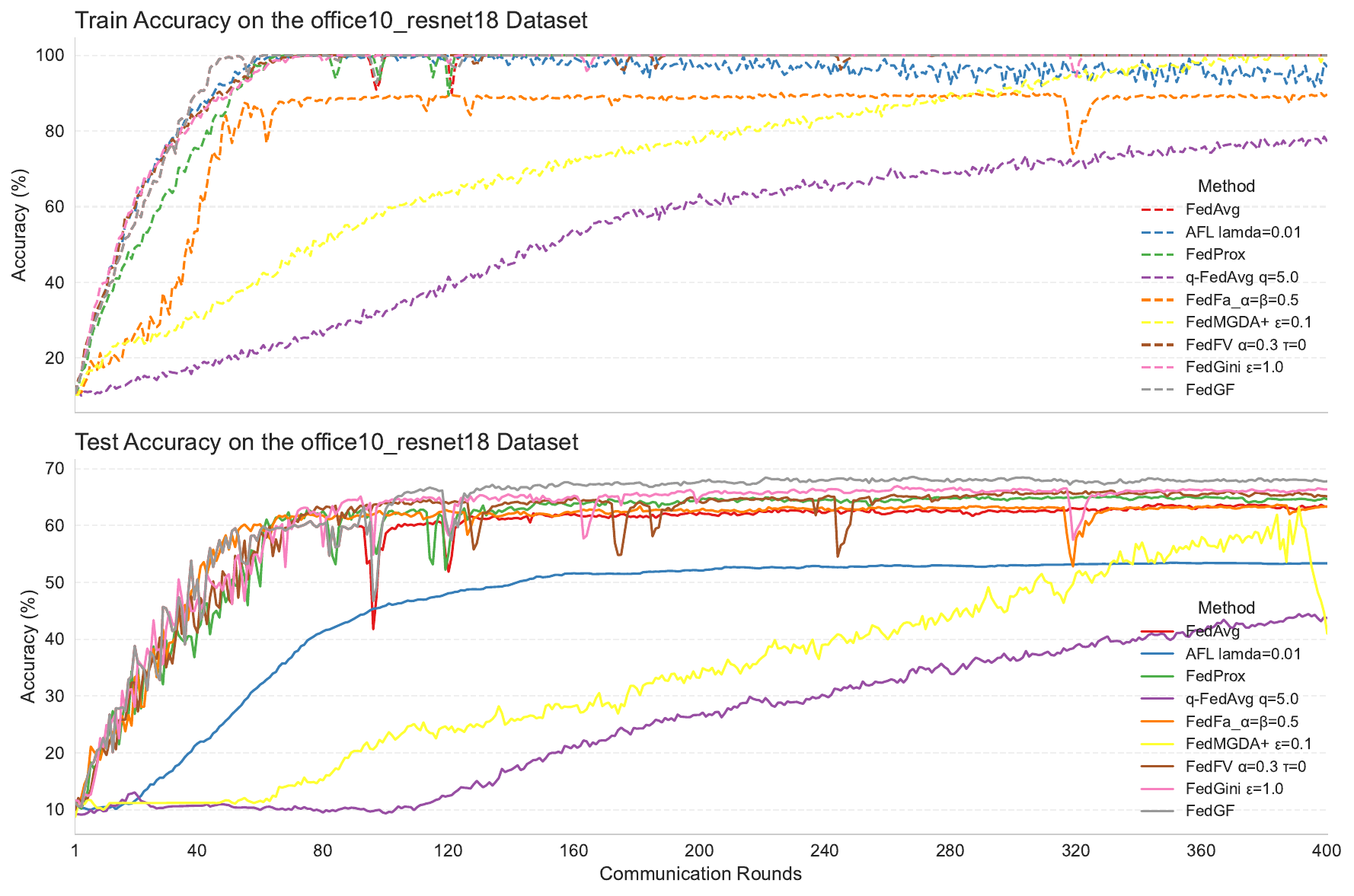}
  \caption{Convergence and Generalization Performance on the Office10 Dataset with ResNet18 network.}
  \Description{}
  \label{fig:office10_train_test}
\end{figure}

\textbf{Figure \ref{fig:office10_train_test}} displays the training and test accuracy trajectories over 400 communication rounds for federated learning methods on the Office10 dataset using ResNet18. FedGA exhibits rapid convergence and strong generalization, reaching over 80\% training accuracy within 40 rounds and maintaining the highest test accuracy above 68\% throughout training. This performance demonstrates both optimization efficiency and robustness to overfitting under heterogeneous client distributions. In contrast, fairness-oriented methods (FedFa, FedFV, FedGini) achieve test accuracies between 62–66\%, showing slower convergence and weaker generalization compared to FedGA. These results highlight FedGA's effectiveness in handling feature heterogeneity while maintaining stable generalization performance.
\begin{table}[htbp]
  \caption{Standard Deviation of Client Accuracy Across Methods on the Office10 Dataset with ResNet18 network}
  \label{tab:office_std}
  \begin{tabular}{lc}
    \toprule
    Method & Std\\
    \midrule
    FedAvg & 10.42±2.72\\
    AFL~|~${\eta _\lambda } = 0.01$ & 11.25±2.37\\
    FedProx & 8.89±1.88\\
    q-FedAvg~|~$q = 0.5$ & 6.80±1.83\\
    FedFa~|~$\beta  = 0.5$ & 8.34±2.48\\
    FedMgda+~|~$\varepsilon  = 0.1$ & 10.51±1.48\\
    FedFV~|~$\alpha  = 0.3,\tau  = 0$ & 10.89±2.17\\
    FedGini~|~$\varepsilon  = 1.0$ & 8.32±1.79\\
    \textbf{FedGA} & \textbf{6.44±0.83}\\
  \bottomrule
\end{tabular}
\end{table}

\textbf{Table \ref{tab:office_std}} reports the standard deviation (Std) of client-level accuracy distributions for various federated learning algorithms on the Office10\_Resnet18 dataset. Consistent with the Gini coefficient results shown in Figure \ref{fig:office10_fairness_performance}, FedGA achieves the lowest accuracy standard deviation. The fact that FedGA outperforms all baselines on both fairness metrics demonstrates its capability in preserving fairness in federated learning systems under feature heterogeneous conditions.
\begin{figure}[htbp]
  \centering
  \includegraphics[width=\linewidth]{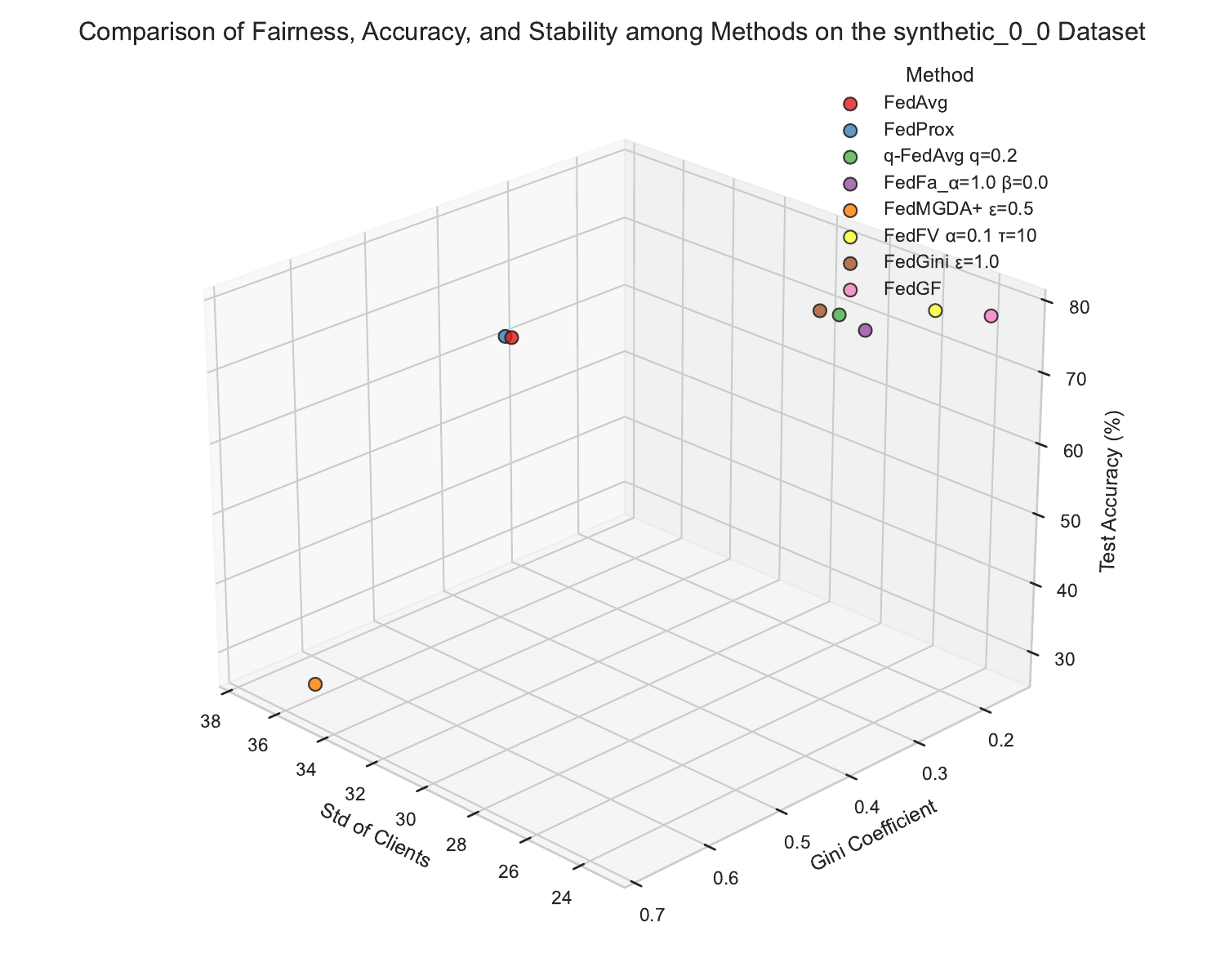}
  \caption{3D Comparison of Performance, Fairness, and Stability on the Synthetic\_0\_0 Dataset.}
  \Description{}
  \label{fig:synthetic_0_0_scatter_3D}
\end{figure}

\textbf{Figure \ref{fig:synthetic_0_0_scatter_3D}} presents a three-dimensional visualization comparing federated learning algorithms across test accuracy, Gini coefficient, and client accuracy standard deviation on the synthetic\_0\_0 dataset. FedGA achieves a favorable balance among these metrics: high accuracy (77.86$\pm$2.24\%), the lowest Gini coefficient (0.16601$\pm$0.02285), and lowest standard deviation (23.38$\pm$2.63). Baseline methods (FedAvg and FedProx) exhibit lower accuracy, higher Gini coefficients, and higher standard deviations, indicating limited resilience to client heterogeneity. While fairness-oriented methods (FedGini and FedFV) show partial improvements in fairness metrics, they remain suboptimal compared to FedGA. The distinct positioning of FedGA in this 3D trade-off space demonstrates its effectiveness in reconciling the competing objectives of accuracy and fairness under heterogeneous conditions.
\begin{figure}[htbp]
  \centering
  \includegraphics[width=\linewidth]{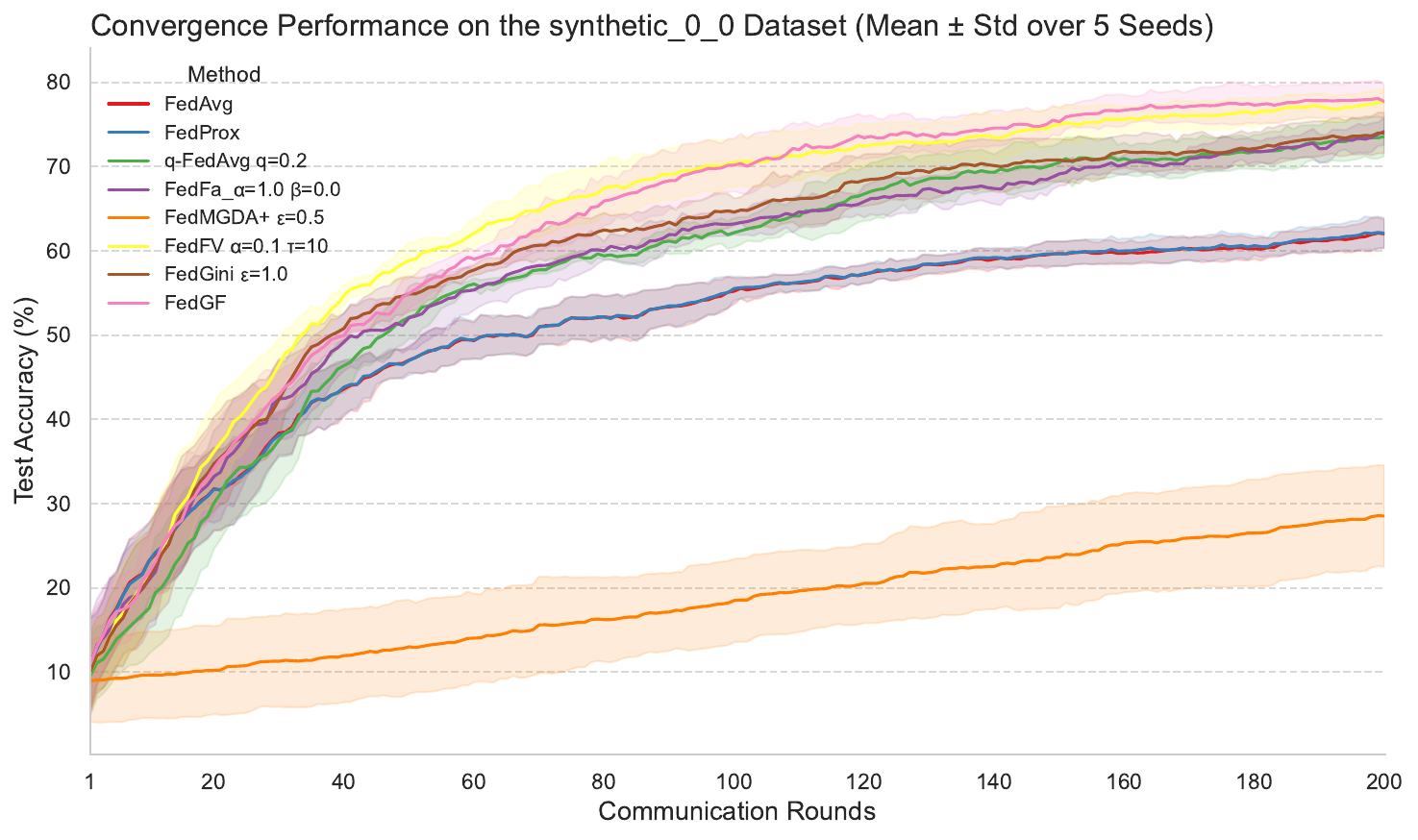}
  \caption{3D Comparison of Performance, Fairness, and Stability on the Synthetic\_0\_0 Dataset.}
  \Description{}
  \label{fig:synthetic_0_0_mean_std_curves}
\end{figure}
\begin{figure}[htbp]
  \centering
  \includegraphics[width=\linewidth]{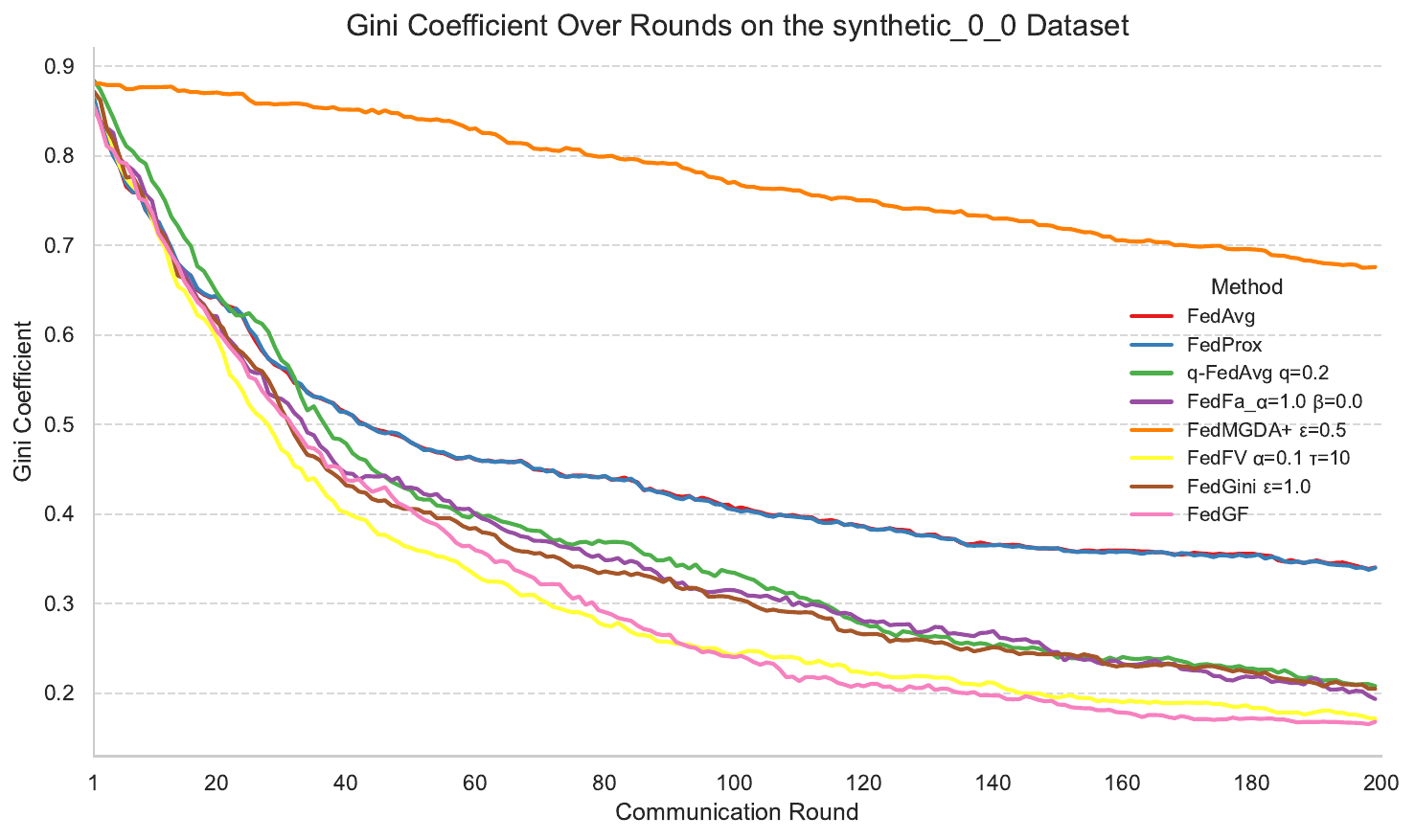}
  \caption{3D Comparison of Performance, Fairness, and Stability on the Synthetic\_0\_0 Dataset.}
  \Description{}
  \label{fig:synthetic_0_0_gini_curve}
\end{figure}

\textbf{Figures \ref{fig:synthetic_0_0_mean_std_curves} and \ref{fig:synthetic_0_0_gini_curve}} present the performance of federated learning algorithms on the synthetic\_0\_0 dataset over 200 rounds, showing mean test accuracy and Gini coefficient evolution, respectively. While FedFV initially converges faster, FedGA achieves the highest final accuracy after ~100 rounds and demonstrates more consistent fairness improvement through sustained Gini coefficient reduction. Other fairness-aware methods (q-FedAvg, FedFa, FedGini) show moderate improvements but remain limited compared to FedGA. FedAvg and FedProx exhibit similar trajectories, reflecting their sensitivity to data heterogeneity.
\begin{table}[htbp]
  \caption{Average Test Accuracy of the Bottom 10\% Clients Under Different FL Algorithms on the synthetic\_0\_0 dataset}
  \label{tab:synthetic_0_0_bottom}
  \begin{tabular}{lc}
    \toprule
    Method & Worst 10\%\\
    \midrule
    FedAvg & 0.00±0.00\\
    FedProx & 0.00±0.00\\
    q-FedAvg~|~$q = 0.2$ & 11.78±8.65\\
    FedFa~|~$\beta  = 0$ & 16.47±6.51\\
    FedMgda+~|~$\varepsilon  = 0.5$ & 0.00±0.00\\
    FedFV~|~$\alpha  = 0.1,\tau  = 10$ & 23.65±8.75\\
    FedGini~|~$\varepsilon  = 1.0$ & 9.90±8.34\\
    \textbf{FedGA} & \textbf{28.94±7.89}\\
  \bottomrule
\end{tabular}
\end{table}

\textbf{Table \ref{tab:synthetic_0_0_bottom}} reports the average test accuracy of the bottom 10\% clients. FedGA achieves the highest bottom-10\% accuracy (28.94$\pm$7.89), demonstrating strong support for disadvantaged clients. While fairness-aware methods (FedFV: 23.65$\pm$8.75, FedFa: 16.47$\pm$6.51, q-FedAvg: 11.78$\pm$8.65) outperform standard baselines, they remain less effective than FedGA. FedAvg, FedProx, and FedMGDA+ yield near-zero accuracy for bottom-performing clients, indicating their focus on global performance at the expense of equity. These results underscore FedGA's effectiveness in addressing the critical challenge of balancing global accuracy with tail fairness in heterogeneous federated learning.
\begin{figure}[htbp]
  \centering
  \includegraphics[width=\linewidth]{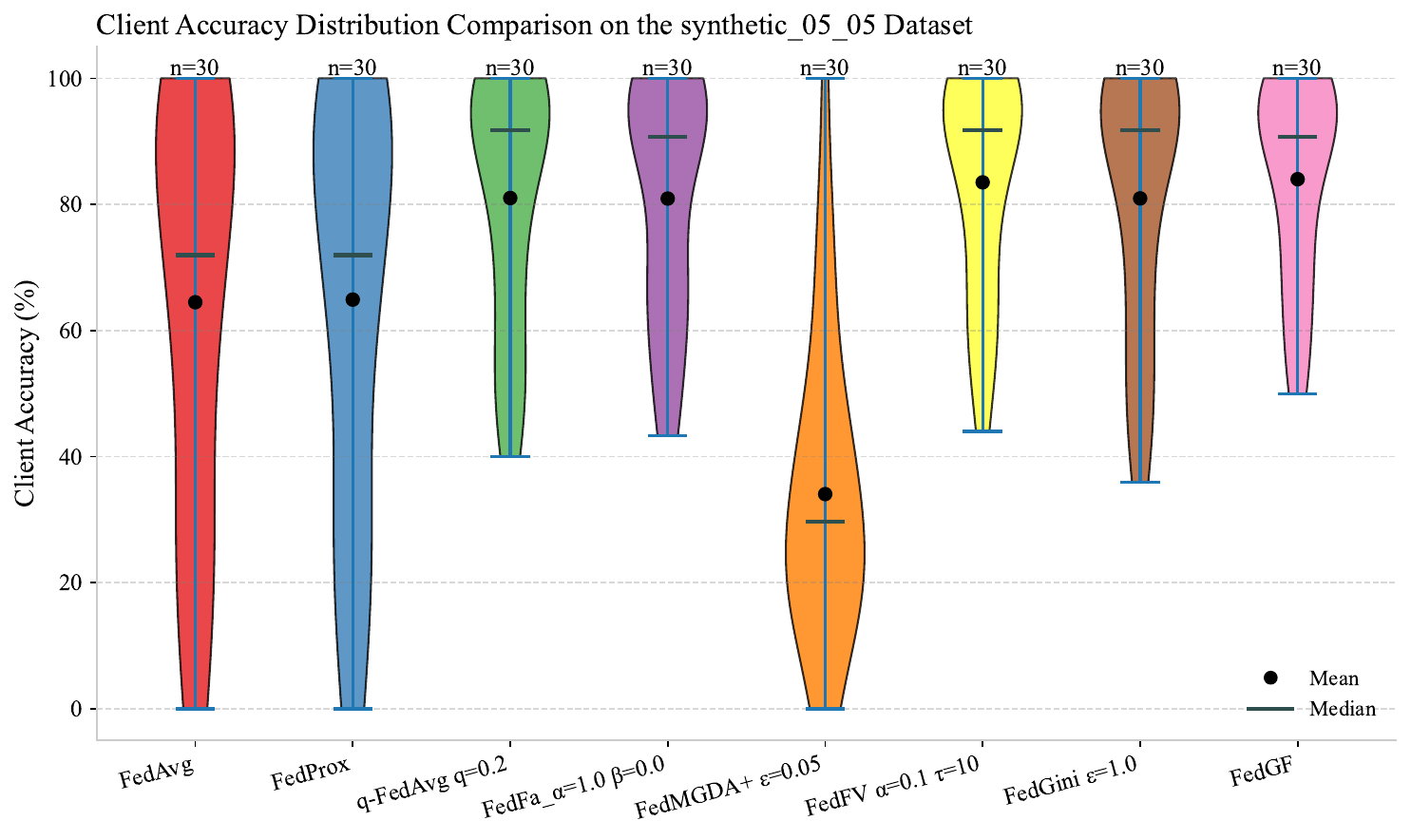}
  \caption{Client Accuracy Distribution Across Algorithms on the Synthetic\_0.5\_0.5 Dataset.}
  \Description{}
  \label{fig:synthetic_05_05_violin}
\end{figure}

\textbf{Figure \ref{fig:synthetic_05_05_violin}} shows violin plots of client-wise accuracy distributions for federated learning algorithms on the synthetic\_05\_05 dataset. FedGA produces a highly concentrated distribution centered near the upper performance range (mean: 84.00$\pm$1.85\%), with its narrow shape and minimal tail mass indicating both high average accuracy and reduced inter-client variability. This demonstrates improved fairness as even disadvantaged clients achieve competitive performance. Standard baselines (FedAvg, FedProx) exhibit broader, bottom-heavy distributions reflecting larger performance gaps across clients. While fairness-oriented methods (FedGini, FedFV) narrow the lower tail and raise median accuracy, their distributions remain wider with lower overall averages than FedGA, highlighting FedGA's effectiveness in balancing global performance with client equity.
\begin{figure}[htbp]
  \centering
  \includegraphics[width=\linewidth]{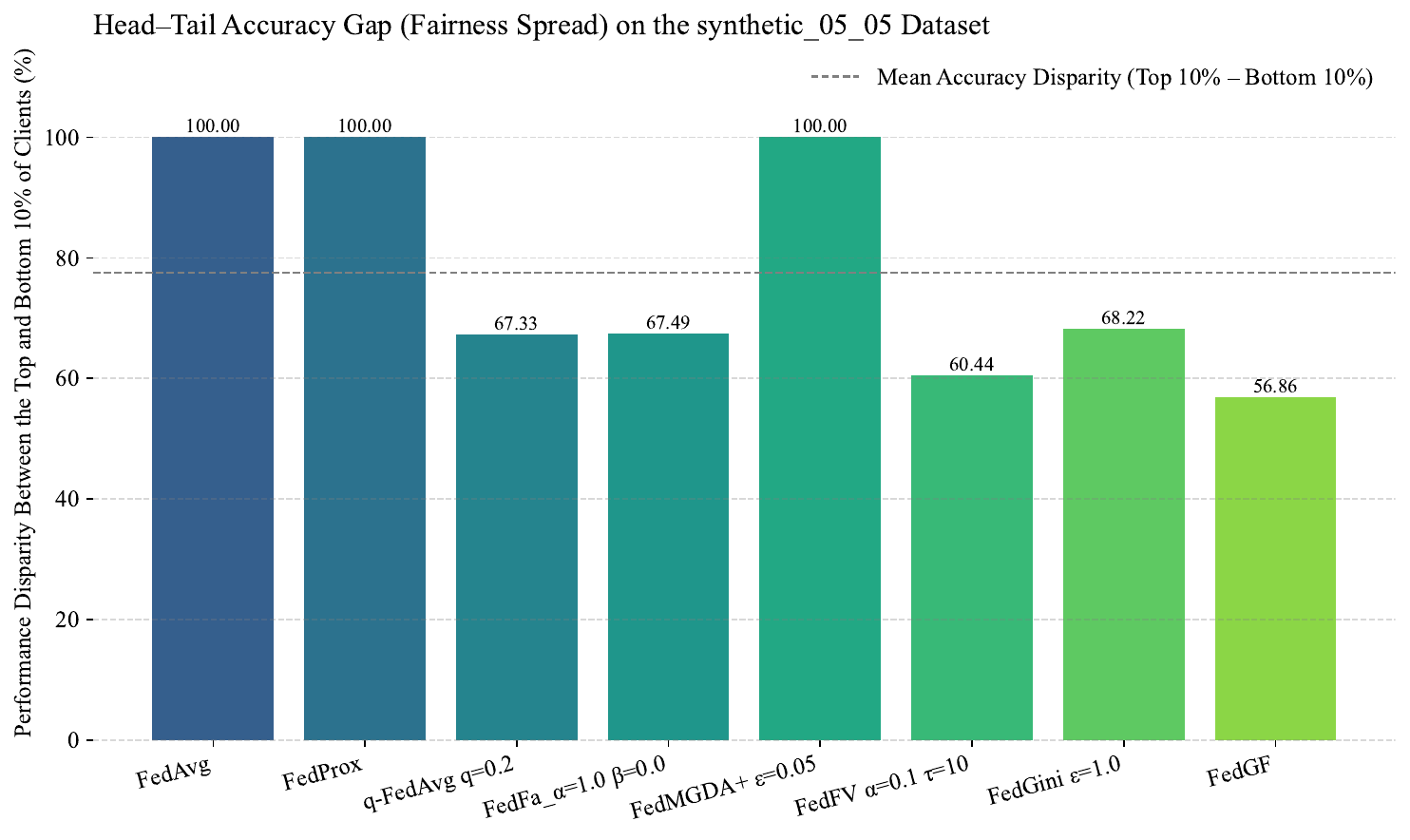}
  \caption{Top–Bottom Client Accuracy Gap Across Algorithms on the Synthetic\_05\_05 Dataset.}
  \Description{}
  \label{fig:synthetic_05_05_head_tail}
\end{figure}

\textbf{Figure \ref{fig:synthetic_05_05_head_tail}} depicts the mean accuracy gap between the top-10\% and bottom-10\% clients on the synthetic\_05\_05 dataset, directly measuring performance polarization. FedGA achieves the smallest gap (56.86\%), indicating that performance improvements are evenly distributed across clients and effectively supporting both well-resourced and disadvantaged participants. Standard approaches (FedAvg, FedProx) exhibit maximum disparities (100\%), reflecting their tendency to favor clients with more representative data. Fairness-oriented algorithms show moderate improvements: q-FedAvg (67.33\%), FedFa (67.49\%), FedFV (60.44\%), and FedGini (68.22\%), demonstrating partial success in reducing head-tail disparities but remaining less effective than FedGA in promoting equitable performance distribution.
\begin{table*}[htbp]
  \caption{Comparison of accuracy and fairness of different methods on the synthetic\_05\_05 dataset}
  \label{tab:synthetic_05_05_acc_fair}
  \begin{tabular}{lcccc}
    \toprule
    Method & Mean acc & Std & Worst 10\% & Best Gini\\
    \midrule
    FedAvg & 64.48$\pm$0.83	& 37.62$\pm$1.56 & 0.00$\pm$0.00 & 0.32495$\pm$0.01057\\
    FedProx & 64.90$\pm$0.60 & 37.64$\pm$1.50 & 0.00$\pm$0.00 & 0.32160$\pm$0.01017\\
    q-FedAvg~|~$q = 0.2$ & 81.01$\pm$1.88 & 22.43$\pm$1.39 & 32.67$\pm$3.27 & 0.15008$\pm$0.01415\\
    FedFa~|~$\beta  = 0$ & 81.25$\pm$1.58 & 22.03$\pm$1.58 & 32.95$\pm$4.01 & 0.14704$\pm$0.01341\\
    FedMgda+~|~$\varepsilon  = 0.05$ & 34.06$\pm$5.87 & 38.68$\pm$2.74 & 0.00$\pm$0.00 & 0.62998$\pm$0.05413\\
    FedFV~|~$\alpha  = 0.1,\tau  = 10$ & 83.50$\pm$1.89 & 19.76$\pm$2.25 & 39.56$\pm$6.98 & 0.12774$\pm$0.01728\\
    FedGini~|~$\varepsilon  = 1.0$ & 80.95$\pm$2.10 & 22.74$\pm$2.13 & 31.78$\pm$4.53	& 0.15121$\pm$0.01827\\
    \textbf{FedGA} & \textbf{84.00$\pm$1.85} & \textbf{18.60$\pm$2.36} & \textbf{43.14$\pm$5.94} & \textbf{0.11955$\pm$0.01837}\\
  \bottomrule
\end{tabular}
\end{table*}
\textbf{Table \ref{tab:synthetic_05_05_acc_fair}} presents a multi-metric evaluation on the Synthetic\_05\_05 dataset. FedGA achieves the highest mean accuracy (84.00±1.85\%) alongside optimal fairness metrics: lowest standard deviation (18.60±2.36), Gini coefficient (0.11955±0.01837), and best worst-10\% client accuracy (43.14±5.94), demonstrating effective balance between utility and equity. Fairness-oriented methods (FedFV, FedGini) show improvements over standard baselines with reduced Gini values (0.12774, 0.15121) and better tail performance, yet remain inferior to FedGA. FedAvg and FedProx exhibit high Gini coefficients and near-zero worst-client accuracy, highlighting their limitations under heterogeneous conditions.
\begin{figure}[htbp]
  \centering
  \includegraphics[width=\linewidth]{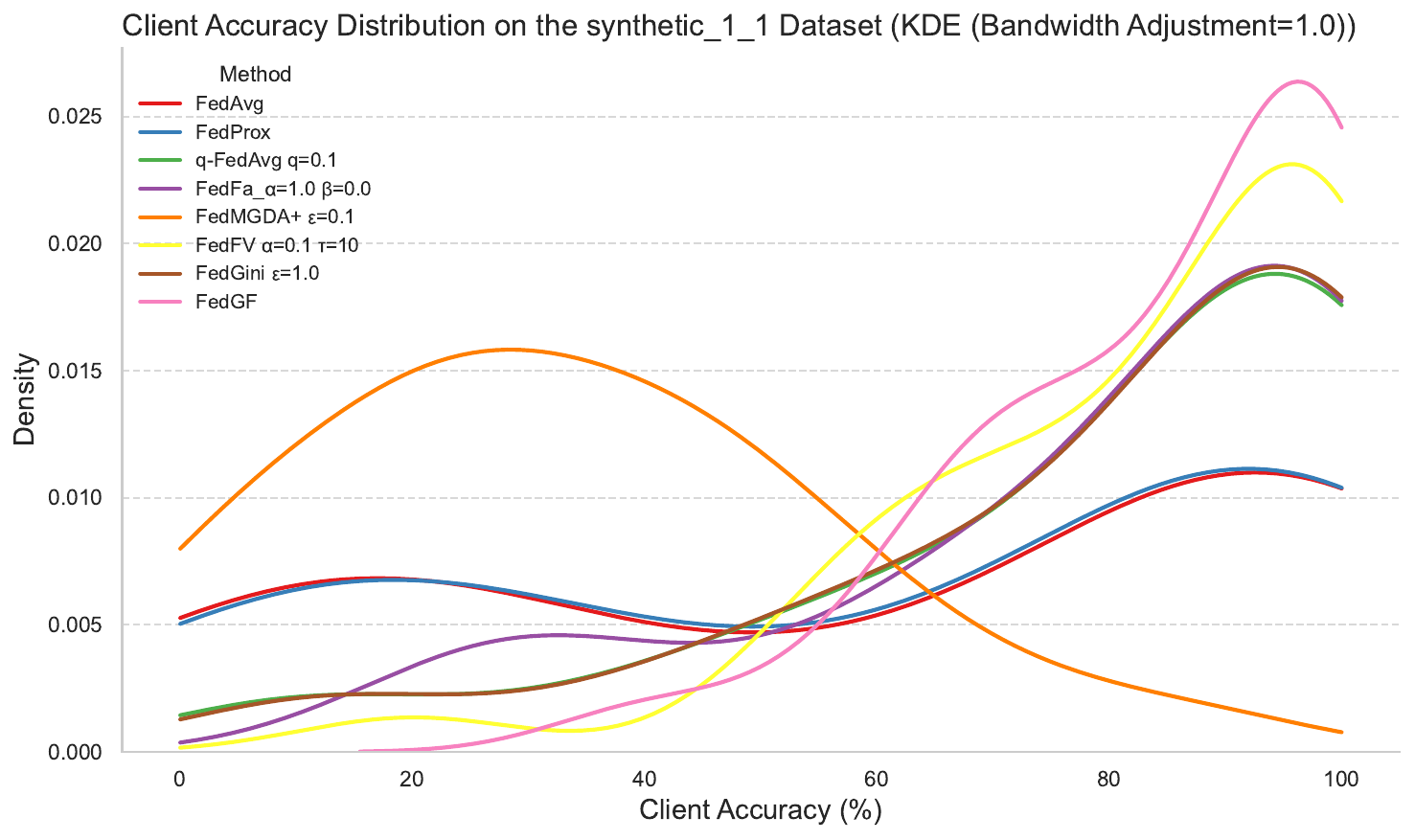}
  \caption{Client Accuracy Distribution via KDE on the Synthetic\_1\_1 Dataset.}
  \Description{}
  \label{fig:synthetic_1_1_accuracy_kde_curves}
\end{figure}

\textbf{Figure \ref{fig:synthetic_1_1_accuracy_kde_curves}} displays kernel density estimates of client test accuracies for federated learning algorithms on the synthetic\_1\_1 dataset. FedGA produces a sharply peaked, right-shifted distribution centered around 85–90\% accuracy with minimal low-accuracy occurrences, reflecting uniformly high performance and low inter-client variability. Conversely, FedAvg and FedProx show broader distributions with substantial density between 40–70\% and pronounced left tails, indicating larger performance disparities. While fairness-enhancing methods (FedFV, FedGini, q-FedAvg, FedFa) shift distributions rightward compared to FedAvg, their curves remain wider and less concentrated than FedGA's, suggesting persistent variability despite fairness improvements.
\begin{figure}[htbp]
  \centering
  \includegraphics[width=\linewidth]{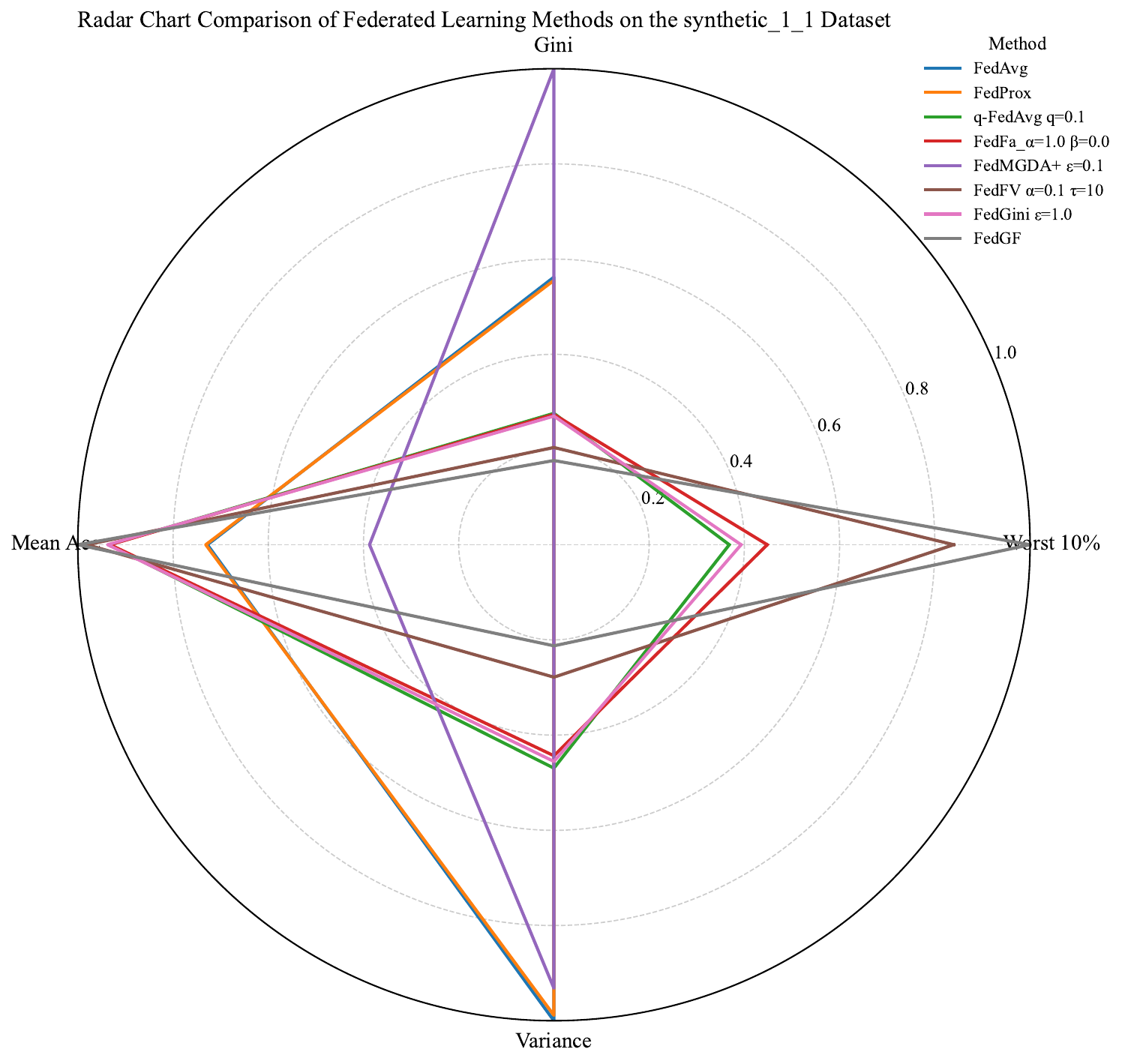}
  \caption{Radar Chart Comparison of Federated Learning Algorithms on Fairness and Performance (Synthetic\_1\_1 Dataset).}
  \Description{}
  \label{fig:synthetic_1_1_radar}
\end{figure}

\textbf{Figure \ref{fig:synthetic_1_1_radar}} illustrates a radar chart comparing federated learning algorithms on the synthetic\_1\_1 dataset across four normalized metrics: mean accuracy, worst 10\% client accuracy, Gini coefficient, and accuracy standard deviation. FedGA exhibits the most balanced performance with the lowest Gini coefficient and highest worst-10\% accuracy, while maintaining competitive mean accuracy and standard deviation. This demonstrates effective fairness improvement without sacrificing overall utility. FedAvg and FedProx achieve moderate mean accuracy but show elevated Gini coefficients and standard deviations, indicating higher disparity. Fairness-oriented methods (q-FedAvg, FedFa, FedGini, FedFV) improve fairness metrics compared to FedAvg but display uneven radar profiles, suggesting trade-offs between different performance dimensions. 
\subsection{Hyperparameter Analysis}
\begin{figure*}[htbp]
  \centering
  \begin{subfigure}{0.32\linewidth}
    \includegraphics[width=\linewidth]{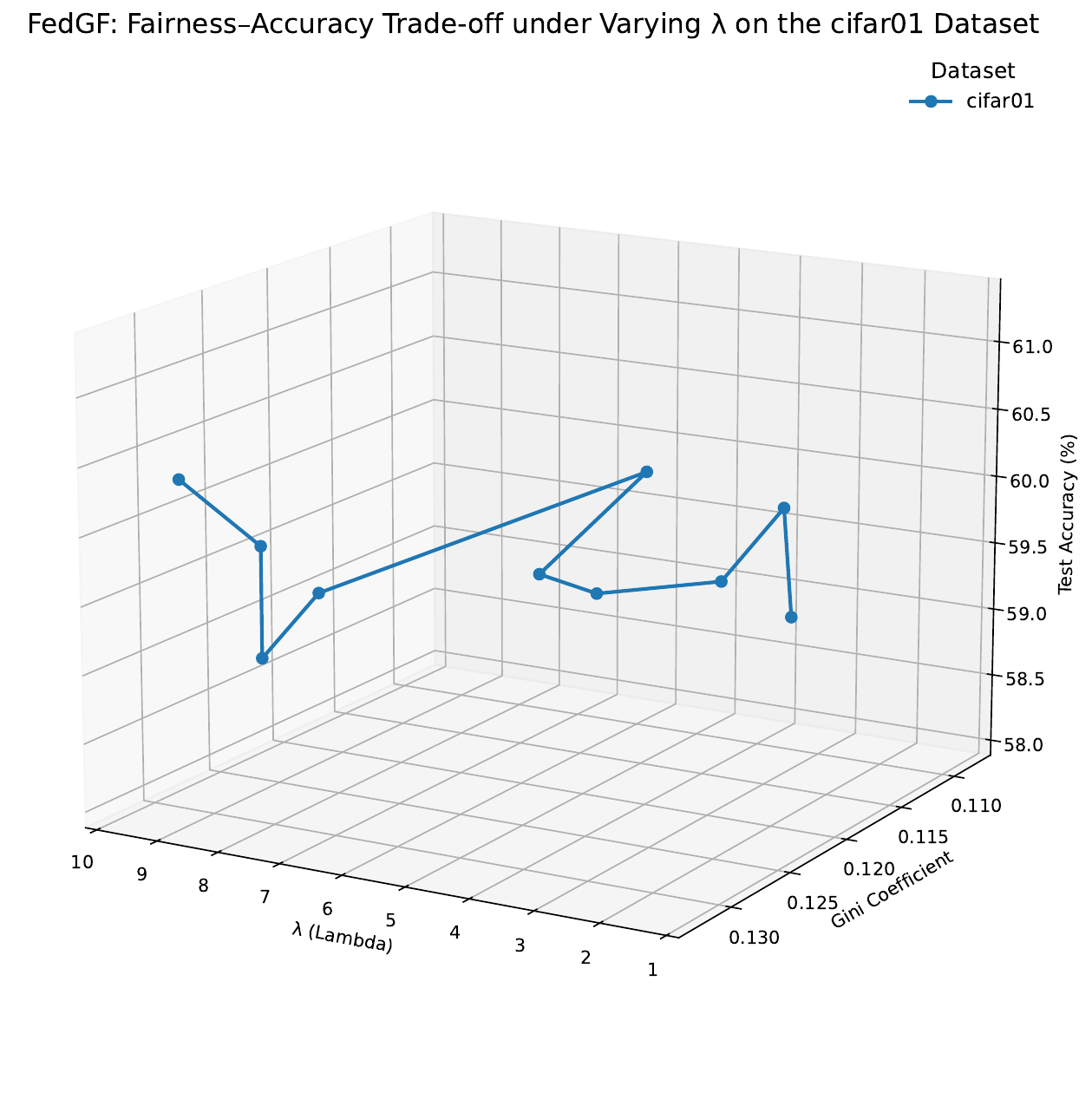}
    \caption{CIFAR01}
  \end{subfigure}
  \begin{subfigure}{0.32\linewidth}
    \includegraphics[width=\linewidth]{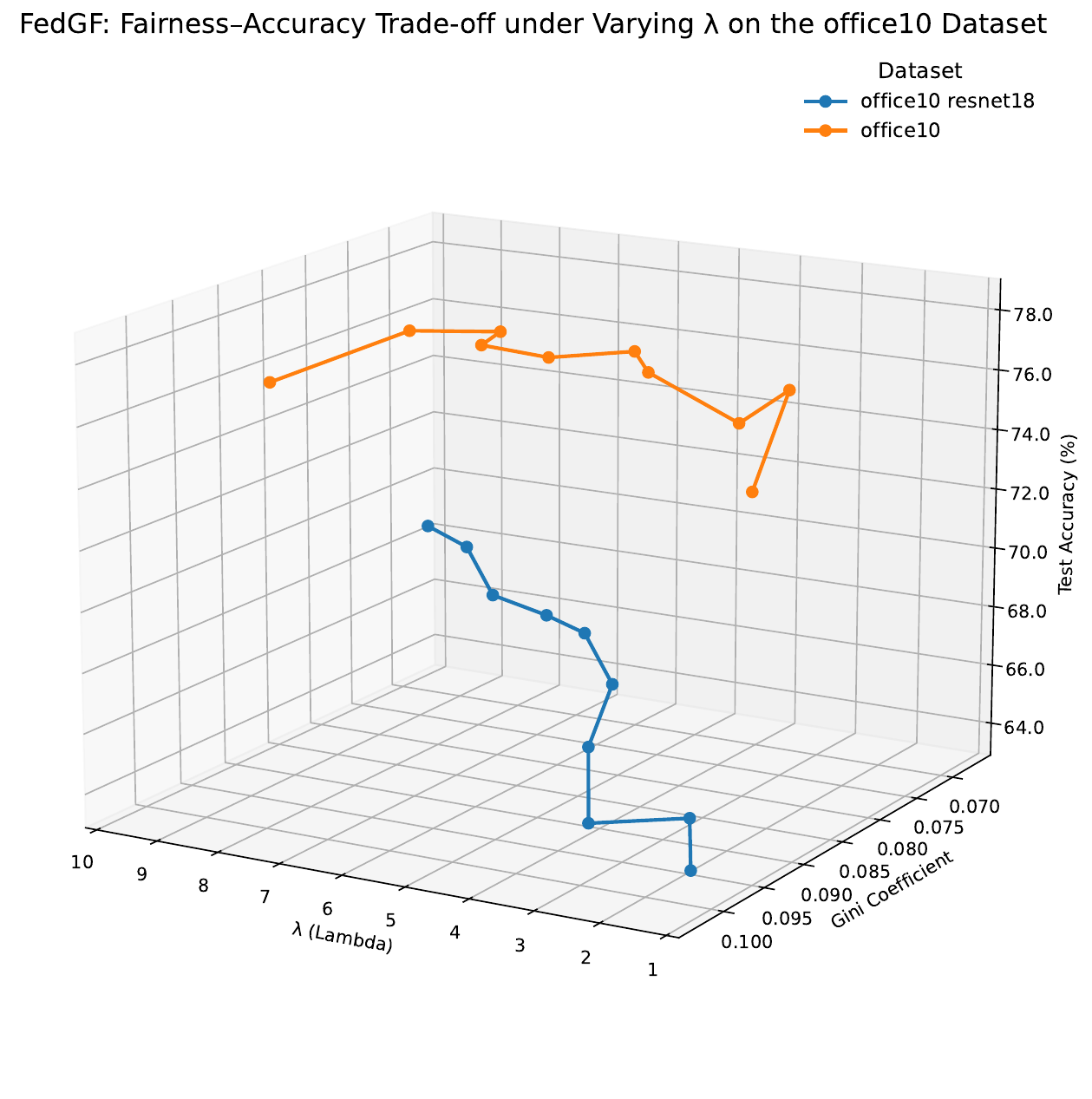}
    \caption{Office10}
  \end{subfigure}
  \begin{subfigure}{0.32\linewidth}
    \includegraphics[width=\linewidth]{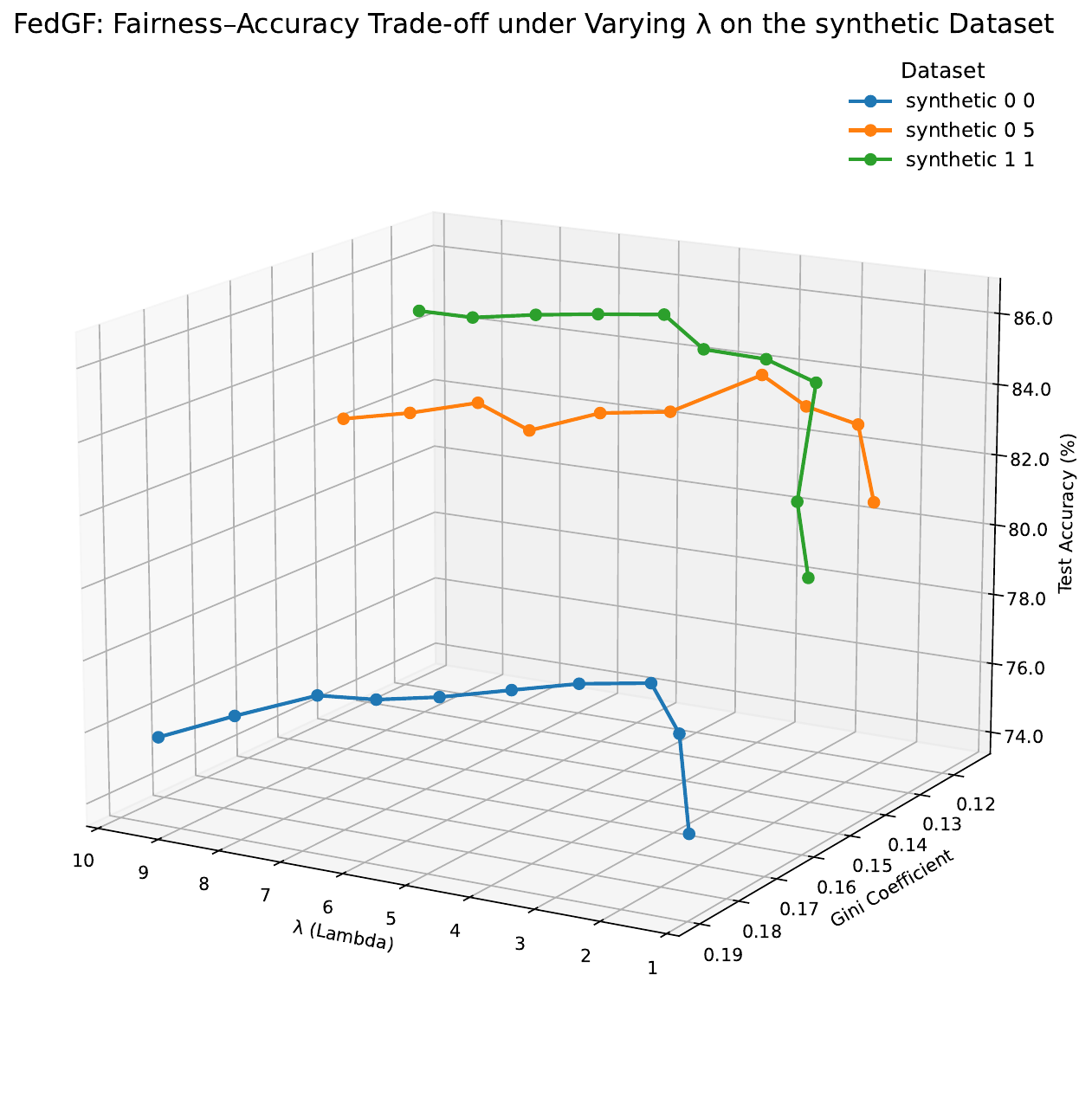}
    \caption{Synthetic}
  \end{subfigure}
  \caption{The impact of hyperparameters on the results.}
  \label{fig:hyperparam}
\end{figure*}

This section examines the impact of hyperparameter $\lambda$ on fairness in federated learning. We evaluated $\lambda$ values from 1 to 10 across datasets, with results shown in Figure 16. The parameter $\lambda$ controls fairness intervention strength—larger values assign higher weights to poorly-performing clients, while $\lambda  = 0$ applies uniform weighting (distinct from FedAvg's data-size-based weighting).\\
As illustrated, most datasets exhibit a U-shaped pattern where the Gini coefficient initially decreases then increases with $\lambda$, indicating that moderate fairness intervention improves equity while excessive intervention may be counterproductive. Office10\_ResNet18 shows a different pattern: after a brief increase at $\lambda  \approx 2$, the Gini coefficient steadily declines, reaching its minimum at $\lambda  = 10$. These dataset-dependent responses demonstrate that optimal $\lambda$ selection is crucial for balancing fairness and utility, as simply increasing $\lambda$ does not guarantee improved fairness and may compromise overall performance.
\subsection{Ablation Experiment}
\begin{table*}[htbp]
  \caption{Ablation experiment on the office-10 dataset}
  \label{tab:ablation_office10}
  \centering
  \begin{tabular}{llccccccc}
    \toprule
    Model & Method & Amazon & Caltech & DSLR & Webcam & Average & Std & Best Gini \\
    \midrule
    \multirow{2}{*}{Alexnet} 
    & FedGA ablation & 78.96$\pm$0.71 & \textbf{67.11$\pm$1.46} & \textbf{73.12$\pm$4.24} & \textbf{88.14$\pm$1.52} & \textbf{76.83$\pm$0.82} & 8.05$\pm$0.61 & 0.07656$\pm$0.00572 \\
    & FedGA          & \textbf{79.38$\pm$0.71} & 66.76$\pm$1.52 & 72.50$\pm$3.64 & 86.10$\pm$1.98 & 76.18$\pm$0.84 & \textbf{7.48$\pm$1.06} & \textbf{0.07209$\pm$0.01006} \\
    \multirow{2}{*}{Resnet18} 
    & FedGA ablation & 68.75$\pm$2.80 & \textbf{58.84$\pm$0.45} & 64.38$\pm$3.19 & \textbf{77.63$\pm$4.72} & 67.40$\pm$1.30 & 7.28$\pm$1.61 & 0.07547$\pm$0.01582 \\
    & FedGA          & \textbf{70.52$\pm$1.70} & 58.40$\pm$0.60 & \textbf{68.75$\pm$2.80} & 75.25$\pm$2.54 & \textbf{68.23$\pm$0.28} & \textbf{6.44$\pm$0.83} & \textbf{0.06673$\pm$0.01113} \\
    \bottomrule
  \end{tabular}
\end{table*}
\begin{table*}[htbp]
  \caption{Ablation experiment on the Cifar-01 and synthetic dataset}
  \label{tab:ablation_cifar01_synthetic}
  \centering
  \begin{tabular}{llccccc}
    \toprule
    Dataset & Method & Average & Std & Worst 10\% & Best Gini \\
    \midrule
    \multirow{2}{*}{Cifar-01}
    & FedGA ablation & 58.85$\pm$1.21 & 13.34$\pm$1.95 & 34.73$\pm$5.43 & 0.12865$\pm$0.01984 \\
    & FedGA          & \textbf{59.80$\pm$1.12} & \textbf{11.54$\pm$1.48} & \textbf{38.85$\pm$3.41} & \textbf{0.10919$\pm$0.01456} \\
    \multirow{2}{*}{Synthetic\_0\_0}
    & FedGA ablation & 77.48$\pm$2.27 & 23.96$\pm$2.58 & 27.26$\pm$6.75 & 0.17092$\pm$0.02336 \\
    & FedGA          & \textbf{77.86$\pm$2.24} & \textbf{23.38$\pm$2.63} & \textbf{28.94$\pm$7.89} & \textbf{0.16601$\pm$0.02285} \\
    \multirow{2}{*}{Synthetic\_05\_05}
    & FedGA ablation & 83.91$\pm$1.88 & 19.16$\pm$2.12 & 40.98$\pm$6.86 & 0.12281$\pm$0.01645 \\
    & FedGA          & \textbf{84.00$\pm$1.85} & \textbf{18.60$\pm$2.36} & \textbf{43.14$\pm$5.94} & \textbf{0.11955$\pm$0.01837} \\
    \multirow{2}{*}{Synthetic\_1\_1}
    & FedGA ablation & 84.81$\pm$1.64 & 18.43$\pm$2.11 & 44.62$\pm$7.99 & 0.11505$\pm$0.01194 \\
    & FedGA          & \textbf{85.01$\pm$1.58} & \textbf{18.19$\pm$2.10} & \textbf{44.76$\pm$6.18} & \textbf{0.11295$\pm$0.01181} \\
    \bottomrule
  \end{tabular}
\end{table*}
To evaluate delayed fairness intervention in FedGA, we conducted ablation experiments comparing FedGA\_ablation (fairness intervention from round one) against original FedGA (fairness intervention after initial phase). \textbf{Tables \ref{tab:ablation_office10} and \ref{tab:ablation_cifar01_synthetic}} present results across three datasets.\\
FedGA consistently outperforms FedGA\_ablation in fairness metrics—achieving lower Gini coefficients and reduced accuracy standard deviation—while maintaining comparable or superior overall accuracy. Improvements are particularly notable on Synthetic and Cifar01 datasets, where FedGA excels across all metrics. Additionally, FedGA yields higher worst-10\% client accuracy, demonstrating stronger protection for disadvantaged participants. These findings validate our design choice of deferring fairness intervention, enabling the model to establish stable optimization before enforcing fairness objectives, thereby enhancing both equity and global knowledge aggregation in heterogeneous settings.
\subsection{Execution Time Analysis}
\begin{table*}[htbp]
  \caption{Execution Time between FedGini and FedGA}
  \label{tab:execution_time}
  \centering
  \begin{tabular}{lccc}
    \toprule
    time/second(s) & Office10\_Alexnet & Office10\_Resnet18 & Cifar01 \\
    \midrule
    FedGini & $4.90\times 10^{-2} \pm 1.32\times 10^{-4}$ & $1.98\times 10^{-1} \pm 3.55\times 10^{-3}$ & $4.02\times 10^{-1} \pm 5.83\times 10^{-3}$ \\
    FedGA   & \textbf{\boldmath $8.32\times 10^{-6} \pm 2.64\times 10^{-7}$} & \textbf{\boldmath $6.25\times 10^{-6} \pm 4.13\times 10^{-7}$} & \textbf{\boldmath $1.20\times 10^{-5} \pm 1.02\times 10^{-6}$} \\
    \bottomrule
  \end{tabular}
\end{table*}

To evaluate computational efficiency, we compared the runtime overhead of FedGA and FedGini's intervention timing algorithms, described in Section 3.1, on Office-10 and CIFAR-10 datasets. As shown in Table 8, FedGA achieves several orders of magnitude improvement in runtime efficiency over FedGini, demonstrating the practicality of its lightweight intervention mechanism for federated training of large-scale models.
\section{Related Work}
Mohri et al. \cite{r29} proposed AFL, which can optimize for any potential target distribution derived from a mixture of client distributions. However, it is limited to scenarios with a relatively small number of participating devices. Li et al. \cite{r3} introduced q-FedAvg, which adjusts fairness via a parameter q, assigning greater weight to clients with higher loss. Hu et al. \cite{r30} developed FedMGDA+, which enhances the fairness of federated learning without compromising the performance of other devices defends against malicious clients. Tian et al. \cite{r4} proposed $alpha$-FedAvg, incorporating Jain’s Index to measure the fairness of federated learning. This method explores the parameter $alpha$ in $\alpha --fairness$ to balance fairness and accuracy. Huang et al. \cite{r9} proposed FedFa, which sets aggregation weights based on training accuracy and participation frequency, and employs dual-momentum to mitigate forgetting. Wang et al. \cite{r26} attributed fairness issues to gradient conflicts and designed a method to alleviate them. Li et al. \cite{r25} proposed FedGini, which uses the Gini coefficient to measure the level of fairness in federated learning, and they introduced a gradient descent-based method to imporove fairness among participants.
\section{Conclusion And Future Work }
We propose FedGA, a federated learning method that monitors Gini coefficient evolution to determine optimal fairness intervention timing and adjusts aggregation weights based on client validation accuracy. Extensive experiments demonstrate that FedGA outperforms existing methods by significantly reducing performance disparities while maintaining competitive accuracy. The delayed fairness intervention strategy proves particularly effective, allowing models to establish stable optimization trajectories before enforcing equity constraints. Future work will extend FedGA to incorporate additional fairness dimensions, including sensitive attribute protection and demographic subgroup equity, further advancing federated learning for socially responsible applications.

\nocite{*}
\bibliographystyle{ACM-Reference-Format}
\bibliography{sample-base}










\end{document}